\documentclass[11pt]{article}

\usepackage[preprint]{acl}

\usepackage{times}
\usepackage{latexsym}
\usepackage{comment}
\usepackage{multirow}
\usepackage{algorithm}
\usepackage{algorithmic}
\usepackage{amsmath}
\usepackage{booktabs}
\usepackage{threeparttable}
\usepackage{tabularx}
\usepackage{array}
\usepackage{makecell}
\usepackage{stfloats}
\usepackage{amssymb}

\usepackage[T1]{fontenc}

\usepackage[utf8]{inputenc}

\usepackage{microtype}

\usepackage{inconsolata}

\usepackage{graphicx}

\usepackage{tikz}
\usetikzlibrary{arrows.meta,positioning}

\title{Layout-Guided Masking for GROBID: Lightweight Structural Gains in Large-Scale Scientific PDF Ingestion
\thanks{Preprint. Under review.}}

\author{Luca Foppiano$^{1}$\thanks{\;Correspondence: \texttt{luca@sciencialab.com}}, \;
  Sana Khamassi$^{1}$, \; Vipul Gupta$^{2}$ \\
  $^{1}$ScienciaLAB, Portugal \qquad
  $^{2}$Helmholtz-Zentrum Hereon, Germany}

\begin{document}
\maketitle
\begin{abstract}
Transforming scholarly PDFs into machine-readable fulltext remains a bottleneck for large-scale information systems.
Recent vision-based parsers improve accuracy, but need GPUs and may introduce noise into the extracted text.
\texttt{GROBID}, a modular font-stream parser running on CPU, is the de-facto standard for structuring scientific articles and underpins several of the largest open scholarly corpora.
We pair it with a lightweight CPU detector localising figure, table, and paratext (header, footer, page number) regions, encoded as typed-area masks whose tokens are routed to \texttt{GROBID}'s specialised models or discarded.
On two PMC corpora, Bioinformatics (1,926 articles) and Materials Science (2,595), scored against JATS with a section-aware structural protocol, our extension improves over plain \texttt{GROBID} on most metrics (NS $+0.025$/$+0.013$; $+0.086$ paragraph recall on Materials Science, $d_z{=}1.08$), and caption-linked figure recovery improves on both corpora.
On the external Table-BRGM benchmark, table detection recovers F1 $0.16 \to 0.94$ and table structure follows (GriTS-Top $0.27 \to 0.78$, below the strongest GPU system).
On body text, against four vision-based systems (\texttt{Docling}, \texttt{MinerU}, \texttt{olmOCR}, \texttt{dots.ocr}), it has the best paragraph precision on both corpora, the best section detection on Materials Science, and a character error rate within 0.004 of the best GPU parser.
End-to-end on CPU, it costs $2.7$--$3.2\times$ less than the cheapest GPU system (\texttt{Docling}) and $10$--$14\times$ less than generative parsers.
\end{abstract}

\section{Introduction}
Extracting text and structure from scientific literature underpins knowledge-graph construction \citep{foppiano2019automatic, jeangirard2019monitoring, lo-etal-2020-s2orc, du2021softcite, foppiano2023automatic, ammar2018construction, orkg}, dense retrieval \citep{karpukhin_dense_2020} and retrieval-augmented generation \citep{guptav, lewis_retrieval-augmented_2021} and parsing errors propagate downstream.
Scholarly PDFs make that hard: multi-column text, multi-panel figures, irregular tables, mathematical notation, and floating captions \citep{agosti_grobid_2009, Zhu2022-kk}. 
A recent generation of vision-based parsers works from the rendered page instead, reading layout from the image and recovering text with OCR or a vision-language model \citep{document_intelligence_survey}; they differ in how much of the page goes to a generative model and how much to dedicated recognisers (\S\ref{sec:related}).

Full-page vision-based parsing needs GPUs at scale, out of reach for many academic groups; autoregressive generation over page images limits throughput \citep{liao2026hsdtrainingfreeaccelerationdocument, li2026efficientdocumentparsingparallel}; and reconstructing text from images rather than from the content stream may introduce transcription errors or hallucination \citep{kydlicek2025finepdfs} in technical content such as gene names, chemical formulas and mathematical notation \citep{thapa2024vision, tong2026does}.

In contrast, \texttt{GROBID} \citep{grobid} is a font-stream parser that reads the PDF content directly, yielding high character-level accuracy on commodity CPUs at scale \citep{foppiano2026scilad}, and is the standard tool in scholarly infrastructure, producing the parsed full-text behind large open corpora and bibliographic services \citep{lo-etal-2020-s2orc, du2021softcite, jeangirard2019monitoring}. 
Its weak point is visually complex pages: figures and tables interrupt the reading order or leak into the surrounding paragraphs \citep{li_figure_2019, soric2025benchmarking}.
    
We take this asymmetry as our starting point: visual processing can be scoped to \emph{localisation} of visually-rich components (e.g. tables, figures) and paratext (or boilerplate, such as page headers and footers). 
We extend \texttt{GROBID} with a lightweight, replaceable object detector that locates figures, tables and paratext.
Its boxes enter the parser as \emph{typed areas}, region extents, which the cascade honours when deciding which model sees which token.
We evaluate the resulting system, \texttt{GROBID+PP}, on 4,521 PubMed Central (PMC) articles from two corpora against JATS-XML, under a perturbation-validated structural protocol with paired significance tests and a component ablation separating masking from routing, alongside four representative vision-based systems \texttt{Docling}, \texttt{MinerU}, \texttt{olmOCR}, and \texttt{dots.ocr} at measured cost.

\section{Related Work}
\label{sec:related}
 
We group the systems discussed here into three families.
Full-page generative parsers, such as \texttt{olmOCR}~\cite{olmocrbench,olmocr2}, \texttt{dots.ocr}~\cite{li2025dotsocrmultilingualdocumentlayout}, and \texttt{FireRed-OCR}~\cite{firredocr} render each page and generate its content autoregressively with a single VLM; full-pipeline document parsers, such as \texttt{Docling}~\cite{livathinos2025doclingefficientopensourcetoolkit}, and \texttt{PP-StructureV3}~\cite{ppstructurev3} combine layout analysis with dedicated OCR, table and formula recognisers; region-routed hybrids, such as \texttt{MinerU 2.5}~\cite{mineru25} and \texttt{GLM-OCR}~\cite{glmocr} apply layout analysis first and route regions to recognisers in parallel (Table~\ref{tab:taxonomy}; Appendix~\ref{app:taxonomy}).
We do not revisit the earlier layout-aware systems \texttt{VILA}~\cite{vila}, \texttt{Nougat}~\cite{nougat} and \texttt{Marker}~\cite{marker}, which the current generation has superseded.

Our extension belongs to the last class: the contribution is orthogonal to the choice of recogniser, showing that the body-text step can be served by a font-stream parser on CPU rather than a VLM, at lower cost.

\section{Layout-Masked GROBID}
\label{sec:pipeline}

\begin{figure*}[t]
\centering
\begin{tikzpicture}[
  font=\footnotesize,
  box/.style={draw, rounded corners=2pt, align=center, minimum height=8mm, inner sep=3pt, text width=#1},
  box/.default=2.5cm,
  data/.style={draw, fill=gray!12, align=center, minimum height=8mm, inner sep=3pt, text width=#1},
  data/.default=2.5cm,
  arr/.style={-{Latex[length=1.8mm]}, thick},
  lab/.style={font=\scriptsize, fill=white, inner sep=1pt, align=center}]
\node[data=1.1cm] (pdf) at (0,0) {PDF};
\node[box] (rast) at (2.9,0.85) {Page\\rasterisation};
\node[box=2.7cm] (det) at (6.3,0.85) {\texttt{PP-DocLayout-L}\\layout detector (CPU)};
\node[data=3.9cm] (areas) at (10.5,0.85) {Figure / table / paratext\\ bounding boxes};
\node[box] (alto) at (2.9,-0.85) {\texttt{pdfalto}\\font-stream tokens};
\node[box=4.4cm] (crf) at (7.0,-0.85) {\texttt{GROBID} fulltext model\\(tokens in typed areas are routed)};
\node[data=2.3cm] (body) at (13.3,-0.85) {TEI \texttt{<body>}};
\node[box=2.9cm] (models) at (9.9,-2.45) {\texttt{GROBID}\\ figure / table models\\(unchanged)};
\node[data=2.3cm] (figs) at (13.3,-2.45) {TEI \texttt{<figure>},\\\texttt{<figure type="table">}};
\node[lab, text=gray] (drop) at (9.9,-3.4) {tokens in \emph{paratext} areas: dropped};
\draw[arr] (pdf) -- (rast.west);
\draw[arr] (pdf) -- (alto.west);
\draw[arr] (rast) -- (det);
\draw[arr] (det) -- (areas);
\draw[arr] (alto) -- (crf);
\draw[arr] (areas.south) -- ++(0,-0.35) -| node[lab, pos=0.25, above] {typed areas} (crf.north);
\draw[arr] (crf) -- node[lab, above] {body tokens} (body);
\draw[arr] (crf.south) |- node[lab, pos=0.60, above] {figure / table\\ tokens (bbox match)} (models.west);
\draw[arr, dashed, gray] (crf.south) |- (drop.west);
\draw[arr] (models) -- (figs);
\draw[arr] (body) -- node[lab, right] {one TEI file} (figs);
\end{tikzpicture}
\caption{Layout-masked \texttt{GROBID}. The detector localises regions, which reach \texttt{GROBID} as typed areas: tokens inside figure/table boxes are withheld from the \emph{fulltext} model and dispatched by bounding-box match to the existing figure and table models, tokens inside paratext boxes are discarded. 
Those models and the TEI writer are identical to plain \texttt{GROBID}'s; only the dispatch differs.}
\label{fig:workflow}
\end{figure*}
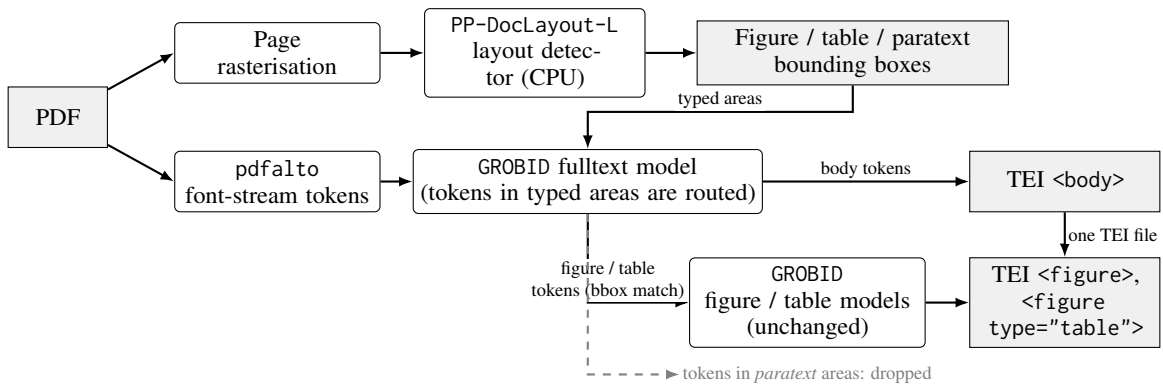

\texttt{GROBID} parses a PDF as a cascade of token-labelling models over the font stream.
\texttt{pdfalto} converts the content stream into layout tokens, each carrying its page, bounding box, and font attributes. A \emph{segmentation} model labels blocks of those tokens into document zones, separating body, and annex from header, bibliography, footnotes, and running headers; a \emph{fulltext} model then labels each token of the body and annex as paragraph, section title, figure, table, equation, list item or reference marker. 
Contiguous spans labelled figure or table go to two specialised models that recover the caption, label, and content of each region. The tokens that remain form the body text, and every stage's output is serialised into one TEI-XML document. 
\texttt{GROBID+PP}, as shown in Figure~\ref{fig:workflow}, changes which tokens the \emph{fulltext} model sees.

\paragraph{Layout detection.}
Each page is rasterised and passed to the PaddlePaddle \texttt{PP-DocLayout} \cite{ppdoclayout} model, a lightweight object detector returning typed bounding boxes. 
We select \texttt{PP-DocLayout-L} on accuracy-to-latency against its S/M variants and \texttt{LADaS} \cite{clericeladas} on the DocLayNet validation set \cite{pfitzmann2022doclaynet}, scoring the five DocLayNet classes that \texttt{GROBID+PP} masks: mean $AP_{50}$ $0.552$ (see Table~\ref{tab:detector-full}) at $0.47$~s/page (Table~\ref{tab:detector-latency}; Appendix~\ref{app:detector}). 
We keep six detector classes: figure, table, caption, running-head, footer and page-number, normalised to PDF page coordinates. 
Page numbers score under DocLayNet's page-footer class, and the last three are the paratext class.
Each figure or table box is assigned to the nearest caption on its page, and all boxes sharing a caption are replaced by their union with it (Appendix~\ref{app:detector}).
This is what turns a multi-panel figure, which the detector reports as one box per panel, back into a single region; a box with no caption nearby is kept on its own.
Three labels reach \texttt{GROBID}: figure, table and paratext; \S\ref{sec:analysis} measures what this merging is worth.

\paragraph{Typed-area masking.}
A \emph{typed area} is a rectangle on a page, labelled \emph{figure}, \emph{table} or \emph{paratext}; the detector produces one list per document, which \texttt{GROBID} takes alongside the PDF through its existing API.
Plain \texttt{GROBID} lets the fulltext model decide which tokens are figure or table content, and typed areas make that decision instead: a token inside a figure or table area skips the fulltext model and goes to the figure or table model with the rest of its area, exactly as if the fulltext model had labelled it so; a token inside a paratext area is dropped; every other token is labelled as before.
Masking changes which model sees a token, not what that model does with it, and the TEI schema is unchanged.

\section{Evaluation}
\subsection{Data and Reference}
\label{sec:data}
We evaluate on two subsets of the PMC Open Access CC-BY corpus: every article ships with a publisher JATS-XML file usable as a body-text reference without manual annotation, and CC-BY lets us redistribute the data with the code.
PMC Bioinformatics corpus comprises 1,943 life-science and medical articles (22,431 pages, 11.55 pages/document)~\cite{constantin2013pdfx}. 
PMC Materials Science comprises 2,595 articles (41,652 pages, 16.05 pages/document) dense in chemical formulas and irregular experimental tables, assembled with ScilitMiner \cite{guptav} from the keywords \emph{materials science}, \emph{physics}, \emph{chemistry}: of 4,000 retrieved documents, 2,595 had a valid PDF and a matching JATS file. 
The two differ in period (84\% of Bioinformatics from 2010--2011, 67\% of Materials Science from 2019--2023): they are contrasting ingestion workloads, not a controlled comparison of domains. 
The corpora are English except for two Bioinformatics articles.

\paragraph{Training-data contamination.} To avoid biasing the results, we resolved both corpora against \texttt{GROBID}'s training data by DOI and by PDF checksum: only 17 Bioinformatics articles overlap and are excluded from every score reported here, for every system, so the comparisons remain paired; Bioinformatics is scored on $n = 1{,}926$.

\paragraph{Ground truth.} 
The reference for the body text is the JATS \texttt{<body>} plus the back matter other than the reference list, footnotes and glossary: \texttt{<ack>}, \texttt{<back>} \texttt{<sec>}, \texttt{<notes>} and appendices are part of the target, as are paragraphs nested in lists and boxes. 
Front matter, the reference list, footnotes, figures, tables, and display equations with their captions are excluded.
Figures and tables are scored on their own. 
Figure recovery is scored against the JATS \texttt{<fig>} captions of the same corpora (\S\ref{sec:figures}). 
Table structure needs cell annotations JATS does not supply, so it is scored on the four datasets of \citet{soric2025benchmarking}, \texttt{PubTables}, \texttt{Table-arXiv}, \texttt{Table-BRGM} and \texttt{ICDAR-2013} (\S\ref{sec:tables}). 
The oracle bound on detector error uses a third PMC corpus of 665 articles for which PubLayNet \citep{publaynet} provides human-annotated figure and table boxes (\S\ref{sec:analysis}).

\subsection{Metrics}
\label{sec:metrics}
All metrics are computed per document and macro-averaged. 
\textbf{NS} ($\uparrow$) is the normalised Levenshtein similarity between reference and predicted body text.
\textbf{WER} ($\downarrow$) and \textbf{CER} ($\downarrow$) are the word and character error rates; reading them together tells whether an error is a misread character or a wrong word boundary.
For structure we match titles and paragraphs one-to-one with the Hungarian algorithm and report \textbf{Sec~F1} for section detection, \textbf{Sec~$\tau$} (Kendall's $\tau_b$) for the order of the matched sections, \textbf{P$\to$S} for the share of matched paragraphs whose predicted section is the match of their gold section, and \textbf{Para~P/R} for paragraph detection.
Because Sec~$\tau$ is computed on matched sections only, a missing section counts against Sec~F1 and not against the order.
Appendix~\ref{app:metrics} gives the definitions. 
Table~\ref{tab:perturb} checks the suite on perturbed copies of the reference: each perturbation moves the one score meant to catch it and leaves the others at identity.

\subsection{Experimental Setup}
\label{sec:setup}

\paragraph{Common representation for fulltext.}
Scoring happens in one representation every format can express, an ordered sequence of sections, each a plain-text title followed by plain-text paragraphs, into which reference and predictions are projected independently.
The elements excluded from the reference in \S\ref{sec:data} are removed from the predictions too, but by different means: structurally for the JATS and for \texttt{GROBID}'s TEI, which mark the body, and by rules for the four vision-based systems, whose JSON or Markdown output carries headings and blocks but no body boundary. 
No converter reads the reference or another system's output (Appendix~\ref{app:projection}).

\paragraph{Hardware.}
\texttt{GROBID} runs on CPU in the \emph{full} deep-learning configuration distributed for production use.
GPU baselines run on \texttt{modal.com}\footnote{\url{https://www.modal.com}} endpoints (versions and concurrency in Table~\ref{tab:cost}): the three generative parsers at one document per A100-40GB, four client workers feeding four containers so that every billed GPU is busy, and \texttt{Docling} (\texttt{docling-serve} 1.7.0, standard pipeline) at sixteen concurrent requests per container on the first available of five GPU types.

\section{Results}
\label{sec:results}
Body text (\S\ref{sec:body}), figures (\S\ref{sec:figures}) and tables (\S\ref{sec:tables}) are scored against the references of \S\ref{sec:data}.
Plain \texttt{GROBID} and \texttt{GROBID+PP} are the same build, differing only in whether typed areas are supplied, so their paired difference is the effect of the extension; the vision-based systems situate it, in what is to our knowledge the first large-scale paired comparison of \texttt{GROBID} with \texttt{Docling}, \texttt{MinerU}, \texttt{olmOCR} and \texttt{dots.ocr} on the same born-digital scientific PDFs under a common protocol.
Table~\ref{tab:fulltext} reports all results from the evaluated systems on both corpora.

\subsection{Body-Text Extraction}
\label{sec:body}

Every system runs on the same document set, paired on the intersection where a baseline returned no parsable output (at most three documents).
Each baseline is compared to \texttt{GROBID+PP} per metric with a paired Wilcoxon signed-rank test, Holm-corrected across system pairs; NS carries a Student-$t$ 95\% confidence interval, and effect sizes are Cohen's $d_z$ on the paired differences.
Comparing \texttt{GROBID+PP} against plain \texttt{GROBID}, gains are significant on every metric on both corpora except section order, at ceiling for both.
On Bioinformatics the gains are led by lexical fidelity and paragraph precision (NS $+0.025$, $d_z{=}0.48$; WER $-0.036$; Para~P $+0.035$); on Materials Science the lexical gain is smaller (NS $+0.013$) but the structural one far larger, paragraph recall $+0.086$ ($d_z{=}1.08$), the largest masked-versus-plain effect in the benchmark.
Improvement is broad: $65\%$ and $56\%$ of documents gain NS against $22\%$ and $11\%$ that lose, and gains above $0.01$ outnumber losses above $0.01$ by $4.5{:}1$ and $8{:}1$.

Against the vision-based systems the picture splits by metric family.
On structure, \texttt{GROBID+PP} has the best paragraph precision on both corpora ($0.885$ and $0.901$; nearest GPU system $0.849$ and $0.879$) and the best section detection on Materials Science ($0.881$ against $0.799$ for \texttt{dots.ocr} and $0.705$ for \texttt{olmOCR}).
At the character level \texttt{GROBID+PP} is at par with the best generative system (CER $0.069$ vs.\ \texttt{MinerU}'s $0.066$; $0.061$ vs.\ \texttt{dots.ocr}'s $0.057$; $d_z{=}0.03$ on both); at the word level the generative parsers pull ahead on Materials Science (\texttt{dots.ocr} $0.059$ WER, \texttt{olmOCR} $0.086$, against $0.105$), while \texttt{MinerU} ties on Bioinformatics (ns).
A WER gap without a CER gap means the residual errors are word boundaries, not characters: spacing around formulas, units and symbols, and de-hyphenation, which a page-rendering model reads from the image and a content-stream parser must infer from glyph positions.
The vision systems otherwise trade paragraph precision for recall, the signature of fragmented output (\texttt{Docling} $0.685$ vs.\ our $0.885$ on Bioinformatics).
\texttt{MinerU 2.5}, which shares our architecture (Table~\ref{tab:taxonomy}), is the most informative baseline: \texttt{GROBID+PP} leads it on section detection and paragraph precision, by a wide margin on Materials Science, while it recovers more paragraphs and places matched ones more accurately.
Routing is common ground; what the comparison isolates is where the routed regions go, to a vision-language model or to sequence labellers over the font stream.
The font-stream route wins on structure and cost (\S\ref{sec:cost}), the visual route on word segmentation of chemical text.

\begin{table*}[t]
\centering
\small
\setlength{\tabcolsep}{3pt}
\caption{Body-text extraction on both corpora.
Best per column and corpus in \textbf{bold}; NS as mean\,$\pm$\,95\% CI.
Superscripts compare each baseline to \texttt{GROBID+PP} on that metric (paired Wilcoxon, Holm-corrected): $^{\ast}$~\texttt{GROBID+PP} better, $^{\ddagger}$~baseline better (both $p<0.05$), $^{\mathrm{ns}}$~not significant.
Sec~$\tau$ excludes documents with fewer than two matched sections (1.6\% / 3.0\% for \texttt{GROBID+PP}, up to 13.5\% for the vision systems on Materials Science).
Inline \LaTeX{} is mapped to Unicode (Appendix~\ref{app:projection}).}
\label{tab:fulltext}
\begin{tabular}{llcccccccc}
\toprule
Corpus & System & NS ($\uparrow$) & WER ($\downarrow$) & CER ($\downarrow$) & Sec F1 & Sec $\tau$ & P$\to$S & Para P & Para R \\
\midrule
\multirow{6}{*}{\makecell[l]{\emph{PMC}\\\emph{Bioinformatics}\\$n = 1{,}926$}}
 & \textbf{\texttt{GROBID+PP} (ours)} & 0.9352\,{\scriptsize$\pm$.004} & 0.0996 & 0.0686 & 0.8717 & 0.9872 & 0.9697 & \textbf{0.8849} & 0.9031 \\
 & \texttt{GROBID} & 0.9097$^{\ast}$\,{\scriptsize$\pm$.005} & 0.1359$^{\ast}$ & 0.1003$^{\ast}$ & 0.8601$^{\ast}$ & 0.9871$^{\mathrm{ns}}$ & 0.9603$^{\ast}$ & 0.8504$^{\ast}$ & 0.8849$^{\ast}$ \\
 & \texttt{Docling} & 0.9034$^{\ast}$\,{\scriptsize$\pm$.008} & 0.1427$^{\ast}$ & 0.1022$^{\ast}$ & 0.8382$^{\ast}$ & 0.9881$^{\mathrm{ns}}$ & 0.9772$^{\ddagger}$ & 0.6851$^{\ast}$ & 0.9317$^{\ddagger}$ \\
 & \texttt{MinerU} & \textbf{0.9416}$^{\ddagger}$\,{\scriptsize$\pm$.004} & \textbf{0.0994}$^{\mathrm{ns}}$ & \textbf{0.0657}$^{\ddagger}$ & 0.8628$^{\mathrm{ns}}$ & 0.9872$^{\ddagger}$ & \textbf{0.9876}$^{\ddagger}$ & 0.8485$^{\ast}$ & \textbf{0.9697}$^{\ddagger}$ \\
 & \texttt{olmOCR} & 0.8988$^{\ast}$\,{\scriptsize$\pm$.005} & 0.1249$^{\ast}$ & 0.1175$^{\ast}$ & 0.8139$^{\ast}$ & \textbf{0.9927}$^{\ddagger}$ & 0.9704$^{\mathrm{ns}}$ & 0.8049$^{\ast}$ & 0.9355$^{\ddagger}$ \\
 & \texttt{dots.ocr} & 0.9100$^{\ast}$\,{\scriptsize$\pm$.006} & 0.1105$^{\ast}$ & 0.1008$^{\ast}$ & \textbf{0.8961}$^{\ddagger}$ & 0.9345$^{\ast}$ & 0.9834$^{\ddagger}$ & 0.6581$^{\ast}$ & 0.9586$^{\ddagger}$ \\
\midrule
\multirow{6}{*}{\makecell[l]{\emph{PMC Materials}\\\emph{Science}\\$n = 2{,}595$}}
 & \textbf{\texttt{GROBID+PP} (ours)} & 0.9392\,{\scriptsize$\pm$.005} & 0.1047 & 0.0614 & \textbf{0.8808} & 0.9837 & 0.9800 & \textbf{0.9008} & 0.9167 \\
 & \texttt{GROBID} & 0.9267$^{\ast}$\,{\scriptsize$\pm$.005} & 0.1213$^{\ast}$ & 0.0754$^{\ast}$ & 0.8686$^{\ast}$ & 0.9837$^{\mathrm{ns}}$ & 0.9708$^{\ast}$ & 0.8754$^{\ast}$ & 0.8308$^{\ast}$ \\
 & \texttt{Docling} & 0.7710$^{\ast}$\,{\scriptsize$\pm$.006} & 0.3714$^{\ast}$ & 0.3195$^{\ast}$ & 0.6839$^{\ast}$ & \textbf{0.9992}$^{\ddagger}$ & 0.9867$^{\ddagger}$ & 0.6331$^{\ast}$ & 0.8735$^{\ast}$ \\
 & \texttt{MinerU} & 0.8692$^{\ast}$\,{\scriptsize$\pm$.005} & 0.1995$^{\ast}$ & 0.1654$^{\ast}$ & 0.7218$^{\ast}$ & 0.9991$^{\ddagger}$ & \textbf{0.9919}$^{\ddagger}$ & 0.7965$^{\ast}$ & 0.8853$^{\ast}$ \\
 & \texttt{olmOCR} & 0.9259$^{\ast}$\,{\scriptsize$\pm$.004} & 0.0855$^{\ddagger}$ & 0.0803$^{\ast}$ & 0.7048$^{\ast}$ & 0.9988$^{\ddagger}$ & 0.9589$^{\ast}$ & 0.8789$^{\ast}$ & 0.9394$^{\ddagger}$ \\
 & \texttt{dots.ocr} & \textbf{0.9502}$^{\ddagger}$\,{\scriptsize$\pm$.004} & \textbf{0.0594}$^{\ddagger}$ & \textbf{0.0570}$^{\ddagger}$ & 0.7985$^{\ast}$ & 0.9956$^{\ddagger}$ & 0.9898$^{\ddagger}$ & 0.7603$^{\ast}$ & \textbf{0.9718}$^{\ddagger}$ \\
\bottomrule
\end{tabular}
\end{table*}

\subsection{Table Extraction}
\label{sec:tables}
We assess table extraction with the benchmark of \citet{soric2025benchmarking} (\S\ref{sec:data}), scored by detection F1, GriTS and TEDS, with cells matched by bounding-box IoU or by text tokens (Appendix~\ref{app:metrics}).
We compare plain \texttt{GROBID}, \texttt{GROBID+PP} and \texttt{Docling}, the benchmark's strongest tool; Table~\ref{tab:tables-body} summarises bbox mode, with the rest in Appendix~\ref{app:tables}.

\texttt{GROBID+PP} improves plain \texttt{GROBID} on every dataset and metric in both matching modes, qualitatively so on BRGM and ICDAR where it essentially fails.
The table model is unchanged; it is simply handed a correctly bounded region (\S\ref{sec:analysis} places this gain in the routing half).
Detection is now uniformly high, bbox F1 $0.926$--$0.975$ against $0.157$--$0.666$ for plain \texttt{GROBID} and above \texttt{Docling}'s on arXiv and BRGM, so region bounding is no longer the limiting factor; \texttt{Docling} keeps the lead on structure quality on three of four datasets.
The deficit that remains is concentrated in cell content (GriTS\textsubscript{Con} $0.02$--$0.16$ lower, TEDS $0.07$--$0.21$) and is produced by the one component the extension leaves untouched, a table model trained on 75 instances from 22 documents; more annotated regions, rather than better boxes, is the natural next lever.
The effect reproduces on a second, independent benchmark: on the scientific-article subset of \texttt{OmniDocBench} \texttt{GROBID+PP} raises TEDS from $31.5$ to $54.7$ (Appendix~\ref{app:omnidocbench}).

\begin{table}[h]
\centering
\small
\setlength{\tabcolsep}{4pt}
\begin{tabular}{llccc}
\toprule
Dataset & System & F1 & G$_{\mathrm{T}}$ & TEDS \\
\midrule
\multirow{3}{*}{PubTables}
 & GROBID+PP & 0.975 & 0.884 & 0.715 \\
 & GROBID    & 0.666 & 0.774 & 0.605 \\
 & Docling   & \textbf{0.988} & \textbf{0.952} & \textbf{0.858} \\
\midrule
\multirow{3}{*}{arXiv}
 & GROBID+PP & \textbf{0.926} & \textbf{0.862} & 0.583 \\
 & GROBID    & 0.430 & 0.781 & 0.535 \\
 & Docling   & 0.897 & 0.794 & \textbf{0.656} \\
\midrule
\multirow{3}{*}{BRGM}
 & GROBID+PP & \textbf{0.937} & 0.777 & 0.602 \\
 & GROBID    & 0.160 & 0.265 & 0.159 \\
 & Docling   & 0.906 & \textbf{0.823} & \textbf{0.739} \\
\midrule
\multirow{3}{*}{ICDAR}
 & GROBID+PP & 0.948 & 0.826 & 0.687 \\
 & GROBID    & 0.157 & 0.592 & 0.482 \\
 & Docling   & \textbf{0.990} & \textbf{0.969} & \textbf{0.895} \\
\bottomrule
\end{tabular}
\caption{Table extraction, bbox cell matching, under the protocol of \citet{soric2025benchmarking}. 
F1 is table detection at IoU~$>0.5$, G$_{\mathrm{T}}$ is GriTS\textsubscript{Top}. Best per dataset in \textbf{bold}. 
Token mode, GriTS\textsubscript{Con} and the token-matching detection rates in Table~\ref{tab:table-full}.}
\label{tab:tables-body}
\end{table}

\subsection{Figure Extraction}
\label{sec:figures}
We evaluate figure extraction comparing plain \texttt{GROBID} and \texttt{GROBID+PP} only.
The JATS gold standard (\S\ref{sec:data}) records no PDF coordinates, so figure \emph{localisation} cannot be scored by IoU; we score \emph{recovery}, matching each gold \texttt{<fig>} caption one-to-one to a TEI \texttt{<figDesc>} at $\theta = 0.70$, tables excluded.
Precision counts every emitted figure, captionless ones included, which can never match a gold caption (Appendix~\ref{app:metrics}).
Caption-anchored region formation (\S\ref{sec:pipeline}) keeps such regions from arising by forming one mask region per logical figure (Appendix~\ref{app:figures}).

\texttt{GROBID+PP} improves caption-linked figure recovery on both corpora, F1 $0.718 \to 0.775$ on Bioinformatics and $0.858 \to 0.885$ on Materials Science, with caption fidelity on matched pairs rising too ($d_z$ $0.41$ and $0.33$, both $p<10^{-50}$; Table~\ref{tab:figures-body}).
The gain is recall at precision parity: \texttt{GROBID+PP} recovers figures plain \texttt{GROBID} misses without emitting proportionally more spurious regions.

\subsection{Ablations of masking and routing}
\label{sec:analysis}
Masking and routing move together in the comparisons above, so we ablated them separately (Appendix~\ref{app:ablation}).
Masking produces the body-text gain in full: routing is worth $+0.0003$ NS on Bioinformatics and $-0.0002$ on Materials Science ($p<0.001$).
That follows from the mechanism (\S\ref{sec:pipeline}): routed regions leave the body stream, so routing can only withhold body text, never add it.
Routing is instead what keeps masking from being destructive: without it, figure recovery collapses (F1 $0.858 \to 0.111$ on Materials Science) and tables emitted fall from $4{,}122$ over $1{,}391$ documents to $17$ over $13$.

Detection quality bounds what masking can achieve, so we re-ran the system with human-annotated boxes in place of the detector's (Appendix~\ref{app:oracle}).
On 665 PubLayNet articles (\S\ref{sec:data}), masking figures and tables raises NS from $0.9204$ to $0.9250$; the human-annotated boxes reach $0.9257$, a further $+0.0007$ (95\% CI $[-0.0004, +0.0022]$).
The detector already collects 87\% of what masking those region types can deliver.
Region formation matters more than box accuracy: the deployed configuration reaches $0.9393$, $+0.0143$ over the matched run and twenty times the remaining headroom, and two further arms attribute $+0.0133$ of it to caption-anchored merging (95\% CI $[+0.0114, +0.0151]$), $+0.0011$ to paratext and nothing to masking the unannotated pages (Appendix~\ref{app:oracle}).

\subsection{Deployment and Cost}
\label{sec:cost}
The operational question for an existing \texttt{GROBID} installation is whether it can be upgraded in place, with plain \texttt{GROBID} as the fallback when the detector is unavailable.

\begin{table*}[!t]
\centering
\footnotesize
\setlength{\tabcolsep}{4pt}
\caption{Cost per document and per million documents, Bioinformatics (LS) / Materials Science (MS). GPU rows: total billed Modal spend $\div$ documents; CPU rows: measured wall-time $\times$ Modal CPU+memory rate (Appendix~\ref{app:cost}).
Concurrency is client threads / backend max inputs (GPU) or workers (CPU); s/doc is throughput, not the billing basis, so a per-hour rate cannot be back-derived from the two columns. $^{\dagger}$From a billed run of 50 documents per corpus at the concurrency shown (\$3.36 in total), one figure for both corpora; the other GPU rows are full-corpus campaigns. $^{\S}$L40S, A100-40GB, A10, L4 or T4 whichever Modal allocated.}
\label{tab:cost}
\begin{tabular}{llccrr}
\toprule
System & Hardware & Concurrency & s/doc (LS / MS) & \$/1M docs LS & \$/1M docs MS \\
\midrule
GROBID+PP (ours) & CPU, 10 cores / 16 GiB & 10 & 5.8 / 8.1 & \textbf{968} & \textbf{1,345} \\
GROBID           & CPU, 10 cores / 16 GiB & 10 & 1.23 / 1.63 & 205 & 272 \\
Docling          & first available$^{\S}$ & 4 / 16 & 7.6 / 7.0 & 3,090 & 3,645 \\
olmOCR           & A100-40GB & 4 / 1 & 11.0 / 13.5 & 11,050 & 12,823 \\
dots.ocr         & A100-40GB & 4 / 1 & 12.5 / 15.0 & 13,300 & 13,971 \\
MinerU           & A100-40GB & 4 / 1 & 10.8 / 13.8 & \multicolumn{2}{c}{33,600$^{\dagger}$} \\
\bottomrule
\end{tabular}
\end{table*}

Table~\ref{tab:cost} reports every system at what it actually cost to run: billed \texttt{modal.com} spend per document for GPU systems, measured wall-time at Modal's published rates for CPU systems (Appendix~\ref{app:cost}). 
\texttt{GROBID+PP} is priced as one system, detector and \texttt{GROBID} together, measured end-to-end over the full corpora (4,538 documents, 64,083 pages, 0 failures) at concurrency 10 on a 12-core Intel Xeon E5-2650 v4 without GPU, with detection at 0.41~s/page.
\texttt{Docling} on the same CPU costs seven to eleven times more than \texttt{GROBID+PP} and more than its own GPU endpoint (Appendix~\ref{app:docling-cpu}).

For an existing \texttt{GROBID} deployment the honest figure is the one against plain \texttt{GROBID}: \texttt{GROBID+PP} costs $4.7$--$5.0\times$ more per document (\$205 $\to$ \$968 and \$272 $\to$ \$1,345 per million), and detection is $81\%$\,/\,$80\%$ of that. 
\texttt{GROBID+PP} still undercuts every GPU system: $3.2\times$\,/\,$2.7\times$ below \texttt{Docling}, whose shared containers and cheaper GPU pool make it the most favourable GPU figure, $10$--$14\times$ below the generative parsers, and $25$--$35\times$ below \texttt{MinerU}, whose vLLM server is the slowest per document ($40$--$53$~s of server time).

\section{Conclusion}
For born-digital scientific PDFs the results support a narrow, practically useful claim, which \texttt{GROBID}+PP instantiates: a font-stream parser, given only the \emph{location} of non-textual regions from a cheap detector, recovers paragraphs, section structure, and table regions it otherwise loses, at CPU cost.
On 4,521 PMC articles \texttt{GROBID}+PP improves section detection and paragraph recovery (up to $+0.09$ recall), lowers error rates, restores the table detection that font-stream parsing loses along with much of the structure that depends on it, and improves caption-linked figure recovery, at $2.7$--$3.2\times$ below the cheapest GPU parser.
The gain sits in structure rather than raw transcription, consistent with the intended mechanism: the extension changes which tokens the text model sees, not how it transcribes them; the strongest generative parsers match it at the character level (CER within $0.004$, $d_z{=}0.03$) and lead only at the word level on chemical text, at ten times the cost. 
Withholding tokens produces the body-text gain; routing them to the figure and table models keeps that withholding from destroying the figures and tables. What decides whether an external box helps is its granularity, one region per logical figure rather than per visual panel (\S\ref{sec:figures}). 
Code, \texttt{GROBID} build, projected references and per-document scores will be released with the paper.

\section*{Limitations}

\paragraph{Structural evaluation scope.} 
The section-aware suite scores the flat section sequence: JATS XML section nesting (hierarchy depth) and the reading order of figures and tables are not evaluated. 
Sec~$\tau$ is undefined for documents with fewer than two matched sections and excludes them, and P$\to$S conditions on matched paragraphs, so it must be read together with paragraph precision/recall. 
Within-section paragraph order is part of the suite but at ceiling ($\geq 0.985$) for every system, so it is reported only in the appendix. 
Because the JATS reference has no coordinates, the figure numbers measure caption recovery and figure-set size, not whether a region is drawn in the right place; a coordinate-bearing reference such as DocLayNet would support an IoU evaluation.


\paragraph{Corpus scope.} Both corpora are born-digital, English, PMC-hosted CC-BY articles with a valid content stream and a matching JATS XML file; the evaluation does not cover scanned pages, non-English text or publishers outside that subset. 
Domain, publisher and era are also confounded across the two: Bioinformatics is publisher-diverse and predominantly from 2010--2011, Materials Science largely single-publisher and predominantly from 2019--2023. 
Layouts of the earlier period are on the whole simpler than current multi-panel, heavily floated ones, which likely flatters every font-stream result on Bioinformatics. 
Differences between the corpora therefore cannot be attributed to scientific domain or layout complexity specifically. 
The paired comparisons within each corpus are unaffected, since every system processes the same documents; what remains open is how far the results transfer to other publishers, periods and collections.

\paragraph{Detector choice and ablations.}
The detector is selected on detection quality over the masked classes and CPU latency (Tables~\ref{tab:detector-latency} and~\ref{tab:detector-full}), not end-to-end: we do not run the S/M/L variants through the full pipeline, so the choice rests on the proxy those tables measure rather than on the fulltext scores the pipeline produces. That comparison may also understate \texttt{LADaS}, which is scored on DocLayNet, a corpus that does not cover the historical documents \texttt{LADaS} targets.
The component ablation (\S\ref{sec:analysis}, Appendix~\ref{app:ablation}) separates paratext masking, figure/table masking and routing on the two main corpora, but region \emph{formation} is isolated only on the PubLayNet corpus (Appendix~\ref{app:oracle}), where caption-anchored merging is worth $+0.013$ NS; its value on the main corpora is not measured. The masking-only arm is likewise a floor rather than a tuned masking-only system: it routes nothing, so what it reports is \texttt{GROBID}'s unaided figure and table detection over a token stream that masking has already depleted.
The oracle bound (Appendix~\ref{app:oracle}) closes the detector question for body text only, on a third corpus and for figure and table masks alone; it inherits PubLayNet's box conventions as the definition of a perfect box, and it says nothing about table \emph{structure}, so how much of the residual gap to \texttt{Docling} on tables is detector error remains open.

\paragraph{Recognition-model capacity.} The \texttt{GROBID+PP} extension changes where region boundaries come from, not how the figure and table models read what is inside them, and those two remain the least-trained stage of the cascade: 108 figure and 75 table instances, from 28 and 22 annotated documents, respectively (against 925 documents for the segmentation model). 
With bounding box detection on the external benchmark now above $0.9$ on every dataset, they are what bounds table-structure quality (\S\ref{sec:tables}). 
Annotating more figure and table regions is therefore the obvious complementary lever, and an orthogonal one: it needs no architectural change and would be scored by the same protocol. 
We did not retrain either model, and this is left for future work. 

\paragraph{Cost figures.} The absolute numbers of \S\ref{sec:cost} are one vendor's on-demand rates on one date, and the two row types rest on different bases (billed spend for GPU systems, measured wall-time priced at the published rate for CPU systems). 
The ordering is more robust than the magnitudes, because it follows from the hardware each system requires rather than from the rate card: the extension needs no accelerator at all, so any pricing under which a GPU-hour costs more than a CPU-hour preserves the direction of the comparison, and what moves with the pricing model is the size of the margin. 
The comparison against plain \texttt{GROBID} is pricing-independent: both sides run on the same CPU instance, so the $4.7$--$5.0\times$ factor is a throughput ratio and carries over to any tariff.

\paragraph{Training-data exposure of the baselines.} 
The decontamination of \S\ref{sec:data} covers \texttt{GROBID} only, whose training set is small and enumerable. 
The vision-based baselines are trained on large page collections assembled from public PDFs, documented at the level of sources rather than documents, and PMC Open Access is a standard source; 67\% of Materials Science is from 2019--2023, inside those collections' time windows. 
We could not test either corpus against those training sets, so some of the evaluated pages may have been seen in training. 
The effect, if any, favours the baselines, not \texttt{GROBID+PP}. 
Those collections also reach well beyond scientific articles, so a scholarly-specialised visual model could close part of the gap; impact on downstream tasks such as retrieval and knowledge-graph construction is left to future work.

\paragraph{Projection asymmetry.} 
Reference and predictions are projected independently and no converter reads the reference (Appendix~\ref{app:projection}), but the two sides are trimmed by different means: front matter and reference list are removed structurally from the JATS and from \texttt{GROBID}'s TEI, and by heading rules from the vision-based systems, whose formats carry no structural boundary. 
A rule that misses leaves the material in the body and is charged to the system; a rule that fires too early removes body text and is also charged to the system. 
The rules were tuned by inspecting system outputs, not scores, but they remain a per-format heuristic that the TEI side does not need.


\bibliography{custom}

@misc{ppdoclayout,
  title={{PP-DocLayout}: A Unified Document Layout Detection Model to Accelerate Large-Scale Data Construction},
  author={Sun, Ting and Cui, Cheng and Du, Yuning and Liu, Yi},
  year={2025},
  eprint={2503.17213},
  archivePrefix={arXiv},
  primaryClass={cs.CV},
  url={https://arxiv.org/abs/2503.17213}
}

@article{guptav,
	title        = {SciLitMiner: An Intelligent System for Scientific Literature Mining and Knowledge Discovery},
	author       = {Gupta, Vipul and Paul, Jonathan David Heaton and Schmitt, Ingo and Pyczak, Florian},
	year         = 2026,
	journal      = {Advanced Intelligent Systems},
	volume       = 8,
	number       = 3,
	pages        = {e202501235},
	doi          = {https://doi.org/10.1002/aisy.202501235},
	url          = {https://advanced.onlinelibrary.wiley.com/doi/abs/10.1002/aisy.202501235},
	eprint       = {https://advanced.onlinelibrary.wiley.com/doi/pdf/10.1002/aisy.202501235}
}

@inproceedings{ammar2018construction,
	title        = {Construction of the Literature Graph in Semantic Scholar},
	author       = {Ammar, Waleed  and Groeneveld, Dirk  and Bhagavatula, Chandra  and Beltagy, Iz  and Crawford, Miles  and Downey, Doug  and Dunkelberger, Jason  and Elgohary, Ahmed  and Feldman, Sergey  and Ha, Vu  and Kinney, Rodney  and Kohlmeier, Sebastian  and Lo, Kyle  and Murray, Tyler  and Ooi, Hsu-Han  and Peters, Matthew  and Power, Joanna  and Skjonsberg, Sam  and Wang, Lucy Lu  and Wilhelm, Chris  and Yuan, Zheng  and van Zuylen, Madeleine  and Etzioni, Oren},
	year         = 2018,
	month        = jun,
	booktitle    = {Proceedings of the 2018 Conference of the North {A}merican Chapter of the Association for Computational Linguistics: Human Language Technologies, Volume 3 (Industry Papers)},
	publisher    = {Association for Computational Linguistics},
	address      = {New Orleans - Louisiana},
	pages        = {84--91},
	doi          = {10.18653/v1/N18-3011},
	url          = {https://aclanthology.org/N18-3011/},
	editor       = {Bangalore, Srinivas  and Chu-Carroll, Jennifer  and Li, Yunyao}
}

@inproceedings{orkg,
	title        = {Open Research Knowledge Graph: Next Generation Infrastructure for Semantic Scholarly Knowledge},
	author       = {Jaradeh, Mohamad Yaser and Oelen, Allard and Farfar, Kheir Eddine and Prinz, Manuel and D'Souza, Jennifer and Kismih\'{o}k, G\'{a}bor and Stocker, Markus and Auer, S\"{o}ren},
	year         = 2019,
	booktitle    = {Proceedings of the 10th International Conference on Knowledge Capture},
	location     = {Marina Del Rey, CA, USA},
	publisher    = {Association for Computing Machinery},
	address      = {New York, NY, USA},
	series       = {K-CAP '19},
	pages        = {243–246},
	doi          = {10.1145/3360901.3364435},
	isbn         = 9781450370080,
	url          = {https://doi.org/10.1145/3360901.3364435},
	numpages     = 4
}

@inproceedings{lo-etal-2020-s2orc,
	title        = {{S}2{ORC}: The Semantic Scholar Open Research Corpus},
	author       = {Lo, Kyle  and Wang, Lucy Lu  and Neumann, Mark  and Kinney, Rodney  and Weld, Daniel},
	year         = 2020,
	month        = jul,
	booktitle    = {Proceedings of the 58th Annual Meeting of the Association for Computational Linguistics},
	publisher    = {Association for Computational Linguistics},
	address      = {Online},
	pages        = {4969--4983},
	doi          = {10.18653/v1/2020.acl-main.447},
	url          = {https://aclanthology.org/2020.acl-main.447/},
	editor       = {Jurafsky, Dan  and Chai, Joyce  and Schluter, Natalie  and Tetreault, Joel}
}

@inproceedings{karpukhin_dense_2020,
	title        = {Dense Passage Retrieval for Open-Domain Question Answering},
	author       = {Karpukhin, Vladimir  and Oguz, Barlas  and Min, Sewon  and Lewis, Patrick  and Wu, Ledell  and Edunov, Sergey  and Chen, Danqi  and Yih, Wen-tau},
	year         = 2020,
	month        = nov,
	booktitle    = {Proceedings of the 2020 Conference on Empirical Methods in Natural Language Processing (EMNLP)},
	publisher    = {Association for Computational Linguistics},
	address      = {Online},
	pages        = {6769--6781},
	doi          = {10.18653/v1/2020.emnlp-main.550},
	url          = {https://aclanthology.org/2020.emnlp-main.550/},
	editor       = {Webber, Bonnie  and Cohn, Trevor  and He, Yulan  and Liu, Yang}
}

@article{lewis_retrieval-augmented_2021,
	title        = {Retrieval-augmented generation for knowledge-intensive nlp tasks},
	author       = {Lewis, Patrick and Perez, Ethan and Piktus, Aleksandra and Petroni, Fabio and Karpukhin, Vladimir and Goyal, Naman and K{\"u}ttler, Heinrich and Lewis, Mike and Yih, Wen-tau and Rockt{\"a}schel, Tim and others},
	year         = 2020,
	journal      = {Advances in neural information processing systems},
	volume       = 33,
	pages        = {9459--9474}
}

@article{Zhu2022-kk,
	title        = {{PDFDataExtractor}: A tool for reading scientific text and interpreting metadata from the typeset literature in the portable document format},
	author       = {Zhu, Miao and Cole, Jacqueline M},
	year         = 2022,
	month        = apr,
	journal      = {J. Chem. Inf. Model.},
	publisher    = {American Chemical Society (ACS)},
	volume       = 62,
	number       = 7,
	pages        = {1633--1643},
	language     = {en}
}

@incollection{agosti_grobid_2009,
	title        = {{GROBID}: {Combining} {Automatic} {Bibliographic} {Data} {Recognition} and {Term} {Extraction} for {Scholarship} {Publications}},
	shorttitle   = {{GROBID}},
	author       = {Lopez, Patrice},
	year         = 2009,
	booktitle    = {Research and {Advanced} {Technology} for {Digital} {Libraries}},
	publisher    = {Springer Berlin Heidelberg},
	address      = {Berlin, Heidelberg},
	volume       = 5714,
	pages        = {473--474},
	doi          = {10.1007/978-3-642-04346-8_62},
	isbn         = 9783642043451,
	url          = {http://link.springer.com/10.1007/978-3-642-04346-8_62},
	urldate      = {2026-04-11},
	isbn_online  = 9783642043468,
	editor       = {Agosti, Maristella and Borbinha, José and Kapidakis, Sarantos and Papatheodorou, Christos and Tsakonas, Giannis}
}

@misc{olmocrbench,
	title        = {{olmOCR: Unlocking Trillions of Tokens in PDFs with Vision Language Models}},
	author       = {Jake Poznanski and Jon Borchardt and Jason Dunkelberger and Regan Huff and Daniel Lin and Aman Rangapur and Christopher Wilhelm and Kyle Lo and Luca Soldaini},
	year         = 2025,
	url          = {https://arxiv.org/abs/2502.18443},
	eprint       = {2502.18443},
	archiveprefix = {arXiv},
	primaryclass = {cs.CL}
}

@misc{olmocr2,
	title        = {olmOCR 2: Unit Test Rewards for Document OCR},
	author       = {Jake Poznanski and Luca Soldaini and Kyle Lo},
	year         = 2025,
	url          = {https://arxiv.org/abs/2510.19817},
	eprint       = {2510.19817},
	archiveprefix = {arXiv},
	primaryclass = {cs.CV}
}

@misc{li2025dotsocrmultilingualdocumentlayout,
	title        = {dots.ocr: Multilingual Document Layout Parsing in a Single Vision-Language Model},
	author       = {Yumeng Li and Guang Yang and Hao Liu and Bowen Wang and Colin Zhang},
	year         = 2025,
	url          = {https://arxiv.org/abs/2512.02498},
	eprint       = {2512.02498},
	archiveprefix = {arXiv},
	primaryclass = {cs.CV}
}

@misc{livathinos2025doclingefficientopensourcetoolkit,
	title        = {Docling: An Efficient Open-Source Toolkit for AI-driven Document Conversion},
	author       = {Nikolaos Livathinos and Christoph Auer and Maksym Lysak and Ahmed Nassar and Michele Dolfi and Panos Vagenas and Cesar Berrospi Ramis and Matteo Omenetti and Kasper Dinkla and Yusik Kim and Shubham Gupta and Rafael Teixeira de Lima and Valery Weber and Lucas Morin and Ingmar Meijer and Viktor Kuropiatnyk and Peter W. J. Staar},
	year         = 2025,
	url          = {https://arxiv.org/abs/2501.17887},
	eprint       = {2501.17887},
	archiveprefix = {arXiv},
	primaryclass = {cs.CL}
}

@article{document_intelligence_survey,
	title        = {Large Language Models in Document Intelligence: A Comprehensive Survey, Recent Advances, Challenges, and Future Trends},
	author       = {Ke, Wenjun and Zheng, Yifan and Li, Yining and Xu, Hengyuan and Nie, Dong and Wang, Peng and He, Yao},
	year         = 2025,
	month        = nov,
	journal      = {ACM Trans. Inf. Syst.},
	publisher    = {Association for Computing Machinery},
	address      = {New York, NY, USA},
	volume       = 44,
	number       = 1,
	doi          = {10.1145/3768156},
	issn         = {1046-8188},
	url          = {https://doi.org/10.1145/3768156},
	issue_date   = {January 2026},
	articleno    = 18,
	numpages     = 64
}

@misc{grobid,
	title        = {GROBID},
	author       = {{GROBID contributors}},
	year         = {2008--2026},
	publisher    = {GitHub},
	note         = {Version 0.9.0. Software Heritage: \href{https://archive.softwareheritage.org/swh:1:dir:dab86b296e3c3216e2241968f0d63b68e8209d3c;origin=https://github.com/kermitt2/grobid;anchor=swh:1:rev:COMMIT}{swh:1:dir:dab86b296e3c3216e2241968f0d63b68e8209d3c}},
	howpublished = {\url{https://github.com/grobidOrg/grobid}}
}

@article{li_figure_2019,
	title        = {Figure and caption extraction from biomedical documents},
	author       = {Li, Pengyuan and Jiang, Xiangying and Shatkay, Hagit},
	year         = 2019,
	month        = nov,
	journal      = {Bioinformatics},
	volume       = 35,
	number       = 21,
	pages        = {4381--4388},
	doi          = {10.1093/bioinformatics/btz228},
	issn         = {1367-4803, 1460-2059},
	url          = {https://academic.oup.com/bioinformatics/article/35/21/4381/5428177},
	urldate      = {2021-08-31},
	language     = {en},
	editor       = {Wren, Jonathan}
}

@misc{liao2026hsdtrainingfreeaccelerationdocument,
	title        = {HSD: Training-Free Acceleration for Document Parsing Vision-Language Model with Hierarchical Speculative Decoding},
	author       = {Wenhui Liao and Hongliang Li and Pengyu Xie and Xinyu Cai and Yufan Shen and Yi Xin and Qi Qin and Shenglong Ye and Tianbin Li and Ming Hu and Junjun He and Yihao Liu and Wenhai Wang and Min Dou and Bin Fu and Botian Shi and Yu Qiao and Lianwen Jin},
	year         = 2026,
	url          = {https://arxiv.org/abs/2602.12957},
	eprint       = {2602.12957},
	archiveprefix = {arXiv},
	primaryclass = {cs.CV}
}

@misc{li2026efficientdocumentparsingparallel,
	title        = {Efficient Document Parsing via Parallel Token Prediction},
	author       = {Lei Li and Ze Zhao and Meng Li and Zhongwang Lun and Yi Yuan and Xingjing Lu and Zheng Wei and Jiang Bian and Zang Li},
	year         = 2026,
	url          = {https://arxiv.org/abs/2603.15206},
	eprint       = {2603.15206},
	archiveprefix = {arXiv},
	primaryclass = {cs.CL}
}

@article{foppiano2023automatic,
	title        = {Automatic extraction of materials and properties from superconductors scientific literature},
	author       = {Foppiano, Luca and Castro, Pedro Baptista and Ortiz Suarez, Pedro and Terashima, Kensei and Takano, Yoshihiko and Ishii, Masashi},
	year         = 2023,
	month        = 1,
	day          = 3,
	journal      = {Science and Technology of Advanced Materials: Methods},
	publisher    = {Informa UK Limited},
	volume       = 3,
	number       = 1,
	doi          = {10.1080/27660400.2022.2153633},
	date         = {2023-01-03}
}

@inproceedings{foppiano2019automatic,
	title        = {Automatic Identification and Normalisation of Physical Measurements in Scientific Literature},
	author       = {Foppiano, Luca and Romary, Laurent and Ishii, Masashi and Tanifuji, Mikiko},
	year         = 2019,
	booktitle    = {Proceedings of the ACM Symposium on Document Engineering 2019},
	location     = {Berlin, Germany},
	publisher    = {Association for Computing Machinery},
	address      = {New York, NY, USA},
	series       = {DocEng '19},
	doi          = {10.1145/3342558.3345411},
	isbn         = 9781450368872,
	url          = {https://doi.org/10.1145/3342558.3345411},
	articleno    = 24,
	numpages     = 4
}

@inproceedings{jeangirard2019monitoring,
	title        = {{Monitoring Open Access at a national level: French case study}},
	author       = {Jeangirard, Eric},
	year         = 2019,
	month        = {Jun},
	booktitle    = {{ELPUB 2019 23rd edition of the International Conference on Electronic Publishing}},
	publisher    = {{ElPub}},
	address      = {Marseille, France},
	volume       = {Academic publishing and digital bibliodiversity},
	doi          = {10.4000/proceedings.elpub.2019.20},
	url          = {https://hal.science/hal-02141819},
	hal_id       = {hal-02141819},
	hal_version  = {v1}
}

@article{du2021softcite,
	title        = {Softcite dataset: A dataset of software mentions in biomedical and economic research publications},
	author       = {Du, Caifan and Cohoon, Johanna and Lopez, Patrice and Howison, James},
	year         = 2021,
	journal      = {Journal of the Association for Information Science and Technology},
	volume       = 72,
	number       = 7,
	pages        = {870--884},
	doi          = {https://doi.org/10.1002/asi.24454},
	url          = {https://asistdl.onlinelibrary.wiley.com/doi/abs/10.1002/asi.24454},
	eprint       = {https://asistdl.onlinelibrary.wiley.com/doi/pdf/10.1002/asi.24454}
}

@inproceedings{thapa2024vision,
	title        = {Vision-Language Models for Biomedical Applications},
	author       = {Thapa, Surendrabikram and Naseem, Usman and Zhou, Luping and Kim, Jinman},
	year         = 2024,
	booktitle    = {Proceedings of the First International Workshop on Vision-Language Models for Biomedical Applications},
	location     = {Melbourne VIC, Australia},
	publisher    = {Association for Computing Machinery},
	address      = {New York, NY, USA},
	series       = {VLM4Bio'24},
	pages        = {1–2},
	doi          = {10.1145/3689096.3690770},
	isbn         = 9798400712074,
	url          = {https://doi.org/10.1145/3689096.3690770},
	numpages     = 2
}

@inproceedings{tong2026does,
	title        = {Does Bigger Mean Better? Comparative Analysis of CNNs and Biomedical Vision-Language Models in Medical Diagnosis},
	author       = {Tong, Ran and Liu, Jiaqi and Wang, Tong and Hu, Xin and Liu, Su and Wang, Lanruo and Xu, Jiexi},
	year         = 2026,
	booktitle    = {2026 International Conference on Artificial Intelligence, Computer, Data Sciences and Applications (ACDSA)},
	volume       = {},
	number       = {},
	pages        = {1--6},
	doi          = {10.1109/ACDSA67686.2026.11467941}
}

@misc{soric2025benchmarking,
	title        = {Benchmarking Table Extraction from Heterogeneous Scientific Extraction Documents},
	author       = {Marijan Soric and C{\'e}cile Gracianne and Ioana Manolescu and Pierre Senellart},
	year         = 2025,
	url          = {https://arxiv.org/abs/2511.16134},
	eprint       = {2511.16134},
	archiveprefix = {arXiv},
	primaryclass = {cs.DB}
}

@misc{foppiano2026scilad,
      title={SciLaD: A Large-Scale, Transparent, Reproducible Dataset for Natural Scientific Language Processing}, 
      author={Luca Foppiano and Sotaro Takeshita and Pedro Ortiz Suarez and Ekaterina Borisova and Raia Abu Ahmad and Malte Ostendorff and Fabio Barth and Julian Moreno-Schneider and Georg Rehm},
      year={2026},
      eprint={2512.11192},
      archivePrefix={arXiv},
      primaryClass={cs.CL},
      url={https://arxiv.org/abs/2512.11192}, 
}

@misc{clericeladas,
    author = {Clérice, Thibault and Janès, Juliette and Scheithauer, Hugo and Bénière, Sarah and Gabay, Simon and Romary, Laurent and Sagot, Benoit and Bougrelle, Roxane},
    title = {{Layout Analysis Dataset with SegmOnto (LADaS)}},
    url = {https://github.com/DEFI-COLaF/LADaS},
    year = {2024}, 
}

@inproceedings{constantin2013pdfx,
  author    = {Constantin, Alexandru and Pettifer, Steve and Voronkov, Andrei},
  title     = {PDFX: fully-automated PDF-to-XML conversion of scientific literature},
  booktitle = {Proceedings of the 2013 ACM Symposium on Document Engineering (DocEng)},
  year      = {2013},
  publisher = {ACM}
}

@misc{mineru,
    title={MinerU: An Open-Source Solution for Precise Document Content Extraction},
    author={Wang, Bin and Xu, Chao and Zhao, Xiaomeng and Ouyang, Linke and Wu, Fan and
            Zhao, Zhiyuan and Xu, Rui and Liu, Kaiwen and Qu, Yuan and Shang, Fukai and
            Zhang, Bo and Wei, Liqun and Sui, Zhihao and Li, Wei and Shi, Botian and
            Qiao, Yu and Lin, Dahua and He, Conghui},
    year={2024},
    eprint={2409.18839},
    archivePrefix={arXiv},
}

@misc{mineru25,
    title         = {MinerU2.5: A Decoupled Vision-Language Model for Efficient High-Resolution Document Parsing},
    author        = {Niu, Junbo and Liu, Zheng and Gu, Zhuangcheng and Wang, Bin and Ouyang, Linke and Zhao, Zhiyuan and others},
    year          = {2025},
    eprint        = {2509.22186},
    archivePrefix = {arXiv},
    primaryClass  = {cs.CV},
}

@misc{glmocr,
    title={GLM-OCR Technical Report},
    author={Duan, Shuaiqi and Xue, Yadong and Wang, Weihan and Su, Zhe and
            Liu, Huan and Yang, Sheng and Gan, Guobing and Wang, Guo and
            Wang, Zihan and Yan, Shengdong and Jin, Dexin and Zhang, Yuxuan and
            Wen, Guohong and Wang, Yanfeng and Zhang, Yutao and Zhang, Xiaohan and
            Hong, Wenyi and Cen, Yukuo and Yin, Da and Chen, Bin and
            Yu, Wenmeng and Gu, Xiaotao and Tang, Jie},
    year={2026},
    eprint={2603.10910},
    archivePrefix={arXiv},
    primaryClass={cs.CV},
    doi={10.48550/arXiv.2603.10910}
}

@misc{firredocr, title={FireRed-OCR Technical Report}, author={Wu, Hao and Lou, Haoran and Li, Xinyue and Zhong, Zuodong and Sun, Zhaojun and Chen, Phellon and Zhou, Xuanhe and Zuo, Kai and Chen, Yibo and Tang, Xu and Hu, Yao and Zhou, Boxiang and Wu, Jian and Wu, Yongji and Yu, Wenxin and Liu, Yingmiao and Huang, Yuhao and Xu, Manjie and Liu, Gang and Ma, Yidong and Sun, Zhichao and Qiao, Changhao}, year={2026}, eprint={2603.01840}, archivePrefix={arXiv}, primaryClass={cs.CV} }

@misc{ppstructurev3,
    title={PP-StructureV3: A Robust Document Analysis System},
    author={{PaddlePaddle Team}},
    year={2023},
    howpublished={\url{https://github.com/PaddlePaddle/PaddleOCR}}
}

@misc{omnidocbench,
    title={OmniDocBench: Benchmarking Diverse PDF Document Parsing with Comprehensive Annotations},
    author={Linke Ouyang and Yuan Qu and Hongbin Zhou and Jiawei Zhu and Rui Zhang and Qunshu Lin and Bin Wang and Zhiyuan Zhao and Man Jiang and Xiaomeng Zhao and Jin Shi and Fan Wu and Pei Chu and Minghao Liu and Zhenxiang Li and Chao Xu and Bo Zhang and Botian Shi and Zhongying Tu and Conghui He},
    year={2024},
    eprint={2412.07626},
    archivePrefix={arXiv},
    primaryClass={cs.CV}
}

@misc{marker,
    title={Marker: Fast, high accuracy PDF to Markdown},
    author={Vik Paruchuri},
    year={2023},
    howpublished={\url{https://github.com/VikParuchuri/marker}}
}

@misc{nougat,
    title={Nougat: Neural Optical Understanding for Academic Documents},
    author={Lukas Blecher and Guillem Cucurull and Thomas Scialom and Robert Stojnic},
    year={2023},
    eprint={2308.13418},
    archivePrefix={arXiv},
    primaryClass={cs.LG}
}

@article{vila,
    title = "{VILA}: Improving Structured Content Extraction from Scientific {PDF}s Using Visual Layout Groups",
    author = "Shen, Zejiang  and
      Lo, Kyle  and
      Wang, Lucy Lu  and
      Kuehl, Bailey  and
      Weld, Daniel S.  and
      Downey, Doug",
    editor = "Roark, Brian  and
      Nenkova, Ani",
    journal = "Transactions of the Association for Computational Linguistics",
    volume = "10",
    year = "2022",
    address = "Cambridge, MA",
    publisher = "MIT Press",
    url = "https://aclanthology.org/2022.tacl-1.22/",
    doi = "10.1162/tacl_a_00466",
    pages = "376--392"
}

@inproceedings{grits,
  author    = {Smock, Brandon and Pesala, Rohith and Abraham, Robin},
  title     = {{GriTS}: Grid Table Similarity Metric for Table Structure Recognition},
  booktitle = {Document Analysis and Recognition -- ICDAR 2023},
  year      = {2023},
  pages     = {535--549},
  publisher = {Springer}
}

@inproceedings{teds,
  author    = {Zhong, Xu and ShafieiBavani, Elaheh and Jimeno Yepes, Antonio},
  title     = {Image-based Table Recognition: Data, Model, and Evaluation},
  booktitle = {Computer Vision -- ECCV 2020},
  year      = {2020},
  pages     = {564--580},
  publisher = {Springer}
}

@misc{kydlicek2025finepdfs,
  title={FinePDFs},
  author = {Hynek Kydl{\'\i}{\v{c}}ek and Guilherme Penedo and Leandro von Werra},
  year={2025},
  publisher={{Hugging Face}},
  journal={Hugging Face repository},
  howpublished={\url{https://huggingface.co/datasets/HuggingFaceFW/finepdfs}}
}

@inproceedings{publaynet,
  author    = {Zhong, Xu and Tang, Jianbin and Jimeno-Yepes, Antonio},
  title     = {{PubLayNet}: Largest Dataset Ever for Document Layout Analysis},
  booktitle = {2019 International Conference on Document Analysis and Recognition (ICDAR)},
  year      = {2019},
  pages     = {1015--1022},
  publisher = {IEEE}
}

@inproceedings{pfitzmann2022doclaynet,
  title={Doclaynet: A large human-annotated dataset for document-layout segmentation},
  author={Pfitzmann, Birgit and Auer, Christoph and Dolfi, Michele and Nassar, Ahmed S and Staar, Peter},
  booktitle={Proceedings of the 28th ACM SIGKDD conference on knowledge discovery and data mining},
  pages={3743--3751},
  year={2022}
}

\clearpage
\section*{Appendix}
\appendix

\section{System Taxonomy}
\label{app:taxonomy}

Table~\ref{tab:taxonomy} places every system discussed in the paper in one of the three families of \S\ref{sec:related}, with its published \texttt{OmniDocBench} score where one exists.

\begin{table*}[t]
\centering
\small
\setlength{\tabcolsep}{6pt}
\begin{tabular}{@{}lllc@{}}
\toprule
System & Class & Version evaluated & OmniDocBench v1.5 \\
\midrule
Ours (GROBID+PP)$^{\ast}$ & region-routed & \texttt{GROBID} 0.9.2-SNAPSHOT with typed areas & n/a \\
GROBID$^{\ast}$           & font-stream   & 0.9.2-SNAPSHOT & n/a \\
Docling$^{\ast}$          & full pipeline & \texttt{docling-serve} 1.7.0, standard pipeline & --- \\
MinerU 2.5$^{\ast}$       & region-routed & \texttt{opendatalab/MinerU2.5-2509-1.2B} & 90.67 \\
olmOCR$^{\ast}$           & full-page gen. & \texttt{allenai/olmOCR-2-7B-1025-FP8} & 81.79 \\
dots.ocr$^{\ast}$         & full-page gen. & \texttt{rednote-hilab/dots.ocr} & 88.41 \\
PP-StructureV3   & full pipeline & --- & 86.73 \\
GLM-OCR          & region-routed & --- & 94.62 \\
FireRed-OCR      & full-page gen. & --- & 92.94 \\
\bottomrule
\end{tabular}
\caption{System taxonomy. $^{\ast}$evaluated in this paper. OmniDocBench~\cite{omnidocbench} v1.5 overall scores are reproduced from the respective reports \cite{mineru25,olmocr2,ppstructurev3,glmocr,firredocr} or from the benchmark's own v1.5 leaderboard (\texttt{dots.ocr}), not re-measured here. \emph{n/a} marks a system the benchmark cannot score rather than one that scores badly: \texttt{OmniDocBench} presents each page in isolation, whereas \texttt{GROBID} labels a document-level token stream (Appendix~\ref{app:omnidocbench}). \emph{---} = no published v1.5 score.}
\label{tab:taxonomy}
\end{table*}

\section{Layout Detector Selection and Region Formation}
\label{app:detector}

\paragraph{Caption-anchored region formation.} Two pre-filters first remove noise, both taken from \texttt{GROBID}'s own vector-graphic box calculator: boxes below a minimum area, and boxes wholly contained in a larger box on the same page. 
Each surviving box is then assigned to the nearest caption box on its page by edge-to-edge distance, intersecting boxes counting as distance zero, and all boxes sharing a caption are merged with that caption into their bounding union, giving one typed area per caption. 
A box whose nearest caption lies beyond a fixed distance is kept as its own region, so uncaptioned artwork is not lost.
\begin{table}[h]
\centering
\small
\setlength{\tabcolsep}{3.5pt}
\begin{tabular}{lc}
\toprule
Model & s/page \\
\midrule
PP-DocLayout-S & \textbf{0.057} \\
PP-DocLayout-M & 0.155 \\
PP-DocLayout-L & 0.47 \\
LADaS & 1.06 \\
\bottomrule
\end{tabular}
\caption{Detector cost on the DocLayNet validation set (6,489 pages): single-worker detection on rendered images, Xeon E5-2650 v4 CPU, excluding rasterisation. Fastest in \textbf{bold}. At the multithreaded operating point used in production, rasterisation\,+\,detection together measure 0.41~s/page for the deployed \texttt{PP-DocLayout-L} (\S\ref{sec:cost}). Detection quality for the same four models is in Table~\ref{tab:detector-full}.}
\label{tab:detector-latency}
\end{table}

Table~\ref{tab:detector-full} gives $AP_{50}$ for all eleven DocLayNet classes, the five \texttt{GROBID+PP} uses and the six it ignores. 
Two patterns are worth noting for anyone reusing these detectors. \texttt{List-item} is $0.000$ for every \texttt{PP-DocLayout} variant: Paddle labels list items as ordinary text, so the class is a taxonomy gap rather than a detection failure, and it is the main reason \texttt{LADaS} closes much of the distance on an all-class average ($0.300$, against $0.280$, $0.334$ and $0.454$ for S, M and L) while trailing on the classes we use. 
Detection quality is also markedly higher on the visual classes (\texttt{Table}, \texttt{Picture}) than on the semantic ones
(\texttt{Title}, \texttt{Footnote}), which is the regime the masking relies on.

\begin{table}[h]
\centering
\small
\setlength{\tabcolsep}{3.5pt}
\begin{tabular}{lcccc}
\toprule
 & \multicolumn{3}{c}{PP-DocLayout} & LADaS \\
\cmidrule(lr){2-4}
Class & S & M & L & \\
\midrule
\multicolumn{5}{l}{\emph{figure / table regions}} \\
Table          & 0.354 & 0.496 & \textbf{0.617} & 0.415 \\
Picture        & 0.520 & 0.551 & \textbf{0.583} & 0.472 \\
Caption        & 0.306 & 0.388 & \textbf{0.591} & 0.462 \\
\midrule
\multicolumn{5}{l}{\emph{paratext}} \\
Page-header    & 0.263 & 0.565 & \textbf{0.653} & 0.017 \\
Page-footer    & \textbf{0.327} & 0.203 & 0.317 & 0.068 \\
\midrule
Mean (masked)  & 0.354 & 0.441 & \textbf{0.552} & 0.287 \\
\midrule[\heavyrulewidth]
\multicolumn{5}{l}{\emph{not used}} \\
Formula        & 0.307 & 0.290 & 0.401 & \textbf{0.462} \\
Text           & 0.475 & 0.518 & \textbf{0.725} & 0.570 \\
Section-header & 0.342 & 0.430 & \textbf{0.688} & 0.326 \\
Footnote       & 0.096 & 0.088 & \textbf{0.251} & 0.151 \\
Title          & 0.090 & 0.142 & \textbf{0.173} & 0.020 \\
List-item      & 0.000 & 0.000 & 0.000 & \textbf{0.342} \\
\bottomrule
\end{tabular}
\caption{Per-class $AP_{50}$ on the DocLayNet validation set (6,489 pages), after mapping each detector's native taxonomy onto DocLayNet's eleven classes. Best per class in \textbf{bold}. 
The first five rows are the classes the pipeline masks and the ones Mean averages; \texttt{Page-footer} also receives the detector's page-number boxes. The remaining six are reported for reference and are not used.}
\label{tab:detector-full}
\end{table}

\section{Oracle-Mask Upper Bound}
\label{app:oracle}

The detector is chosen on detection quality (Appendix~\ref{app:detector}), which leaves open how much of the pipeline's residual error is \emph{its} error. 
We bound that directly by replacing the detector with human-annotated boxes and re-running the pipeline unchanged.

\paragraph{Corpus.} PubLayNet \citep{publaynet} annotates figure and table regions on renders of PMC articles, so ground-truth boxes and the JATS gold text used throughout this paper exist for the same document. 
Its validation split is page-sampled (1.44 pages per document) and cannot support a document-level oracle, so we draw from the train split. 
We fetched 2,896 candidate articles from the PMC Cloud service and kept the 665 whose annotations still align with the PDF served today: at least 80\% of pages annotated, and at least 90\% of PDF words falling inside some annotated region. 
Of the 2,231 rejected, 2,060 failed only the page-coverage test, 110 only the word test and 61 both. The word-test failures are cases where the deposited PDF has been replaced since PubLayNet was built in 2019; the page-coverage failures are documents PubLayNet annotated only in part.
This corpus is disjoint from the two evaluation corpora of \S\ref{sec:data}, so the absolute scores below are not comparable to Table~\ref{tab:fulltext}; only the differences between rows are.

\paragraph{Conditions.} Four runs in the first block and four in the second, one \texttt{GROBID} build, one evaluator, 665 documents.
The unmasked run is issued through the same typed-area path with an empty area list, so the conditions differ only in mask content. 
The first three are strictly matched: they mask the same categories (figure and table, the only ones PubLayNet annotates) on the same pages (PubLayNet annotates a prefix of each document), so \emph{detector} and \emph{oracle} differ only in box quality. 
The fourth is the deployed configuration of \S\ref{sec:pipeline}, which differs from the matched run in three ways: every page rather than the annotated prefix, caption-anchored unions rather than raw boxes, and paratext. 
It is not comparable to the oracle and is reported to separate the value of mask \emph{scope} from that of box accuracy. 
Two intermediate arms then separate the three: figure/table boxes on every page without merging, and caption-merged figure/table regions without paratext. 
They were run with the matched and deployed arms on one later build of the same branch, so their rows are comparable with each other (the matched arm reproduces to $0.0001$ NS) but not digit for digit with the first block.

\begin{table}[h]
\centering
\small
\setlength{\tabcolsep}{4pt}
\begin{tabular}{lccc}
\toprule
Masks & NS & F1 & Prec. \\
\midrule
none & 0.9204 & 0.9260 & 0.9126 \\
\texttt{PP-DocLayout-L}, matched & 0.9250 & 0.9320 & 0.9229 \\
PubLayNet ground truth & 0.9257 & 0.9325 & 0.9243 \\
\midrule
\texttt{PP-DocLayout-L}, deployed & \textbf{0.9393} & \textbf{0.9355} & \textbf{0.9298} \\
\midrule[\heavyrulewidth]
\multicolumn{4}{l}{\emph{scope decomposition}} \\
matched & 0.9249 & 0.9282 & 0.9187 \\
+ every page & 0.9249 & 0.9282 & 0.9187 \\
+ caption-anchored merging & 0.9381 & 0.9312 & 0.9250 \\
+ paratext (= deployed) & 0.9392 & 0.9316 & 0.9255 \\
\bottomrule
\end{tabular}
\caption{Oracle-mask experiment, 665 PubLayNet/PMC articles. 
Recall is flat ($0.9395$--$0.9435$) and is omitted. Upper block: the first three rows are matched in category and page scope; the fourth is the deployed configuration. Lower block: the four scope arms on one later build, each row adding one change to the previous.}
\label{tab:oracle}
\end{table}

\paragraph{Result.} Table~\ref{tab:oracle} gives the aggregates and Table~\ref{tab:oracle-ci} the paired bootstrap (10,000 resamples) over per-document differences. 
A perfect figure/table detector is worth $+0.0007$ NS with the confidence interval capping the headroom at $+0.0022$: this is a bounded null rather than an absent effect. \texttt{PP-DocLayout-L} captures 87\% of the oracle's NS gain, 88\% of its precision gain and 92\% of its F1 gain.
Widening what is masked is worth twenty times more than perfecting the boxes, and the decomposition says which widening. Masking the unannotated pages is worth nothing: they add 32 boxes to the matched 4,618, and 656 of 665 projected texts are byte-identical. Caption-anchored merging carries the gain, $+0.0133$ NS with 70\% of documents improving and 9\% worsening: the union of panels plus caption withholds the inter-panel text and the caption from the fulltext model, which the raw boxes leave in the body. Paratext adds $+0.0011$, consistent with the $+0.0022$ and $+0.0002$ the component ablation measures on the main corpora (Appendix~\ref{app:ablation}).

\begin{table}[h]
\centering
\small
\setlength{\tabcolsep}{4pt}
\begin{tabular}{lcc}
\toprule
Contrast & $\Delta$NS & 95\% CI \\
\midrule
none $\to$ detector & $+0.0047$ & $[+0.0017, +0.0077]$ \\
detector $\to$ oracle & $+0.0007$ & $[-0.0004, +0.0022]$ \\
matched $\to$ deployed & $+0.0143$ & $[+0.0124, +0.0162]$ \\
\midrule
matched $\to$ every page & $-0.0000$ & $[-0.0001, +0.0000]$ \\
every page $\to$ merged & $+0.0133$ & $[+0.0114, +0.0151]$ \\
merged $\to$ deployed & $+0.0011$ & $[+0.0007, +0.0015]$ \\
\bottomrule
\end{tabular}
\caption{Paired bootstrap, 10,000 resamples, mean per-document difference in NS ($n = 665$). 
The second row is the oracle headroom; the last three decompose the third.}
\label{tab:oracle-ci}
\end{table}

\paragraph{Dispersion.} Under the \emph{matched} masks the mean improves but the median document does not.
Median $\Delta$NS is $0.0000$; 173 documents gain more than $0.01$ and 188 lose more than $0.01$, and the positive mean rests on a heavy right tail ($p_{95} = +0.086$, $p_{99} = +0.154$, against $p_{5} = -0.041$). 
A Wilcoxon signed-rank test on NS is not significant here ($p = 0.61$) although the paired mean difference is. 
Narrow scope is what produces that spread: the deployed configuration, on the same 665 documents, has a positive median ($+0.0025$), gains on $54\%$ against $27\%$ that lose, and is significant at $p < 10^{-25}$. 
The matched arm isolates box quality and is not the configuration we propose; its dispersion qualifies the oracle comparison above, not the deployed system of \S\ref{sec:body}.

\section{Component Ablation}
\label{app:ablation}

\S\ref{sec:body} compares \texttt{GROBID+PP} against plain \texttt{GROBID}, in which masking and routing move together. We separate them.

\paragraph{Design.} A typed area labelled \emph{paratext} withholds its tokens from the fulltext model and dispatches them nowhere, whereas \emph{figure} and \emph{table} withhold and then route (\S\ref{sec:pipeline}). 
Relabelling the mask payload therefore ablates routing while holding the withheld token set exactly fixed, with no change to the code. 
Four arms, one build, one configuration, one evaluator, the same documents; arms differ only in the payload:

\begin{table}[h]
\centering
\small
\setlength{\tabcolsep}{4pt}
\begin{tabular}{llc}
\toprule
Arm & Typed areas sent & Step \\
\midrule
A & none & plain \\
P & paratext & A$\to$P: paratext mask \\
B & + fig/tab, not routed & P$\to$B: fig/tab mask \\
C & + fig/tab, routed & B$\to$C: routing \\
\bottomrule
\end{tabular}
\caption{Ablation arms; each row adds to the one above. A is plain \texttt{GROBID} and C the deployed extension, so A$\to$C is the comparison of \S\ref{sec:body}.}
\label{tab:ablation-arms}
\end{table}

Masks come from the same caption-anchored \texttt{PP-DocLayout-L} output used throughout, so all three masking arms withhold identical token sets. 

\paragraph{Control.} Arms A and C are plain \texttt{GROBID} and the deployed extension; they reproduce the four rows of Table~\ref{tab:figures-body} on all 16 figure cells, and the ablation is read as paired differences between arms under one build, one configuration and one evaluator. 
The intermediate arms therefore measure the same quantity the headline does.

\begin{table}[h]
\centering
\small
\setlength{\tabcolsep}{3pt}
\begin{tabular}{llccccc}
\toprule
 & Arm & NS & WER & Sec F1 & Para P & Para R \\
\midrule
\multirow{4}{*}{\rotatebox{90}{LS}}
 & A & 0.9097 & 0.1359 & 0.8601 & 0.8504 & 0.8849 \\
 & P & 0.9119 & 0.1333 & 0.8657 & 0.8593 & 0.8852 \\
 & B & 0.9349 & 0.0999 & 0.8717 & 0.8842 & 0.9035 \\
 & C & 0.9352 & 0.0996 & 0.8717 & 0.8849 & 0.9031 \\
\midrule
\multirow{4}{*}{\rotatebox{90}{MS}}
 & A & 0.9267 & 0.1213 & 0.8686 & 0.8754 & 0.8308 \\
 & P & 0.9269 & 0.1212 & 0.8693 & 0.8760 & 0.8307 \\
 & B & 0.9394 & 0.1046 & 0.8808 & 0.9007 & 0.9169 \\
 & C & 0.9392 & 0.1047 & 0.8808 & 0.9008 & 0.9167 \\
\bottomrule
\end{tabular}
\caption{Body text by arm. 
Section order is at ceiling in every arm ($0.9837$--$0.9872$) and is omitted.}
\label{tab:ablation-body}
\end{table}

\begin{table}[h]
\centering
\small
\setlength{\tabcolsep}{3pt}
\begin{tabular}{llccc}
\toprule
 & Step & $\Delta$NS & $d_z$ & Impr. \\
\midrule
\multirow{4}{*}{\rotatebox{90}{LS}}
 & paratext mask & $+0.0022$ & $0.09$ & 23.8\% \\
 & fig/tab mask & $+0.0230$ & $0.48$ & 60.5\% \\
 & routing & $+0.0003$ & $0.08$ & 6.6\% \\
 & full (A$\to$C) & $+0.0254$ & $0.48$ & 64.7\% \\
\midrule
\multirow{4}{*}{\rotatebox{90}{MS}}
 & paratext mask & $+0.0002$ & $0.05$ & 1.4\% \\
 & fig/tab mask & $+0.0125$ & $0.42$ & 55.5\% \\
 & routing & $-0.0002$ & $-0.07$ & 0.8\% \\
 & full (A$\to$C) & $+0.0125$ & $0.41$ & 55.7\% \\
\bottomrule
\end{tabular}
\caption{NS decomposition. Paired Wilcoxon, Holm-corrected within metric; all steps significant at $p < 0.05$ except where noted in the text. \emph{Impr.} is the fraction of documents improving on that step.}
\label{tab:ablation-ns}
\end{table}

\paragraph{Masking produces the body-text gain.} Routing contributes $+0.0003$ NS on Bioinformatics and $-0.0002$ on Materials Science (Table~\ref{tab:ablation-ns}); on paragraph recall it is negative on both ($-0.0004$, $-0.0002$, both $p < 0.001$). 
Routed regions leave the body stream, so routing can only withhold body text, never add it; on Materials Science arm B alone ($0.9394$) edges past the deployed system ($0.9392$). 
The two masking terms are unequal and split by corpus: paratext masking is worth ten times more on Bioinformatics than on Materials Science, while figure/table masking dominates both. 
The largest single effect in the ablation is Materials Science paragraph recall under figure/table masking, $+0.0862$ at $d_z = 1.09$ for the masking step alone; the full plain-to-masked contrast of \S\ref{sec:body} gives $d_z = 1.08$.

\paragraph{Routing prevents masking from being destructive.} Withholding figure and table tokens without routing them removes the evidence \texttt{GROBID}'s own figure and table models depend on
(Table~\ref{tab:ablation-fig}). Emitted \texttt{<figure type="table">} elements fall from $2{,}912$ to $254$ on Bioinformatics and from $4{,}122$ to $17$ on Materials Science, the latter leaving only 13 of $1{,}391$ documents with any table at all; the deployed arm raises both above plain \texttt{GROBID} ($4{,}441$ and $5{,}406$). 
Arm B buys its body-text gain by deleting the document's figures and tables.

\begin{table}[h]
\centering
\small
\setlength{\tabcolsep}{3pt}
\begin{tabular}{llcccc}
\toprule
 & Arm & P & R & F1 & Tables \\
\midrule
\multirow{4}{*}{\rotatebox{90}{LS}}
 & A & 0.711 & 0.726 & 0.718 & 2,912 \\
 & P & 0.731 & 0.730 & 0.731 & 2,896 \\
 & B & 0.177 & 0.104 & 0.131 & 254 \\
 & C & 0.709 & 0.855 & 0.775 & 4,441 \\
\midrule
\multirow{4}{*}{\rotatebox{90}{MS}}
 & A & 0.799 & 0.927 & 0.858 & 4,122 \\
 & P & 0.799 & 0.927 & 0.858 & 4,125 \\
 & B & 0.144 & 0.091 & 0.111 & 17 \\
 & C & 0.813 & 0.971 & 0.885 & 5,406 \\
\bottomrule
\end{tabular}
\caption{Figure recovery at $\theta = 0.70$ under the denominator of \S\ref{sec:figures}, and emitted \texttt{<figure type="table">} count.}
\label{tab:ablation-fig}
\end{table}

\section{Reference and Prediction Projection}
\label{app:projection}

\S\ref{sec:setup} projects every system's output and the JATS reference into one common representation. The rules are as follows.

\paragraph{Structure.} Each document becomes an ordered sequence of sections, each a plain-text title (serialised as a \texttt{\#\#} line) and an ordered list of plain-text paragraphs. 
JATS \texttt{<sec>/<title>} and TEI \texttt{<div>/<head>} become section titles and their \texttt{<p>} descendants become paragraphs, including paragraphs nested in lists, quotations and boxes; for Markdown/JSON output the corresponding native heading and text-block boundaries are used, and one block is one paragraph. 
Nested sections are traversed in document order and flattened, because hierarchy depth is not scored, while a paragraph remains attached to the nearest preceding heading supplied by that source. 
Text before the first heading is retained in an untitled root section. 
The projection preserves the order and boundaries emitted by each system: it neither splits nor merges paragraphs to improve a match, and it does not repair a system's labelling. 
A reference list a system emitted as body text, a running header it promoted to a heading, or an appendix it filed under \emph{References} is scored as emitted.

\paragraph{Filtering.} Reference and predictions are projected independently, and each converter reads only its own output, never the reference or another system's output, so the conversion cannot use gold structure. 
On the reference side, front matter, the reference list, footnotes, the glossary, and figures, tables and display equations with their captions and labels are removed structurally; \texttt{<ack>} keeps its own title, and Open Access licence statements filed under it are dropped. 
On the prediction side, figure and table content is excluded through whatever structure each format supplies: an ancestor test on TEI \texttt{<figure>}, \texttt{<table>} and \texttt{<formula>} for \texttt{GROBID}, block type for \texttt{MinerU}, block category for \texttt{dots.ocr} (which also drops page headers and footers), block label for \texttt{Docling}, and, for \texttt{olmOCR}'s Markdown, HTML \texttt{<table>}/\texttt{<figure>} blocks and caption blocks opening with a figure or table label followed by a capitalised word. 
The reference list is removed when the system placed it under a \emph{References}-like heading. 
Full-page parsers also emit the front matter, which the reference does not contain; it is removed by a rule that reads only the system's headings: the text before an \emph{Introduction}-like heading among the leading headings; otherwise the \emph{Abstract} or \emph{Keywords} section, ending at its first paragraph without a structured-abstract label so that an unheaded introduction survives; otherwise, for papers that open without such headings, everything before the first run of body text, discarding a title-like leading heading. Sidebar badges promoted to headings (\emph{Open Access}, \emph{Check for updates}, \emph{Received}/\emph{Accepted}) are dropped. The rule that fired is recorded per document. 
\texttt{GROBID} needs neither step because its TEI marks the body structurally. 
For \texttt{dots.ocr}, whose deployment had concatenated per-page output in lexicographic order, the page order is restored from the bounding boxes before projection.

\paragraph{Normalisation.} Markup is discarded without inserting spaces, so \texttt{$\alpha$<sub>2</sub>-integrin} stays \emph{$\alpha$2-integrin} on both sides; Markdown emphasis is stripped; inline \LaTeX{} (\texttt{MinerU}, \texttt{olmOCR}, \texttt{Docling}, \texttt{dots.ocr}, and the reference's Springer \texttt{<tex-math>}) is mapped to Unicode with one normaliser. 
Finally, the same string normalisation is applied to every retained title and paragraph: Unicode NFKC, lower-casing, removal of bracketed citation markers, whitespace collapsing and removal of leading section numbers. 
The lexical metrics concatenate the resulting blocks in order; the structural metrics use the retained section and paragraph boundaries. 
We manually inspected projected samples from every system in both corpora, including nested sections, figure/table-adjacent text and the worst-scoring documents of each system; the converters, the per-document rule logs and the projected references are released with the code.

\section{Metric Definitions and Validation}
\label{app:metrics}
All metrics are computed on the normalised body text of one document (reference $a$, prediction $b$) and macro-averaged over documents.

\paragraph{Lexical.} For reference $a$ and prediction $b$ as normalised character strings, with $\mathrm{Lev}$ the character-level Levenshtein distance (\texttt{rapidfuzz} implementation),
\begin{equation}
  \mathrm{NS}(a, b) = 1 - \frac{\mathrm{Lev}(a, b)}{\max(|a|, |b|)} .
  \label{eq:ns}
\end{equation}
Writing $a_w, b_w$ for the same texts as word sequences,
\begin{equation}
  \mathrm{WER}(a,b) = \min\!\left(1, \frac{\mathrm{Lev}(a_w, b_w)}{|a_w|}\right) ,
  \label{eq:wer}
\end{equation}
\begin{equation}
  \mathrm{CER}(a,b) = \min\!\left(1, \frac{\mathrm{Lev}(a, b)}{|a|}\right) .
  \label{eq:cer}
\end{equation}

\paragraph{Matching.} All structural scores rest on one primitive. Given gold items $X = (x_1 \dots x_m)$, predicted items $\hat{X} = (\hat{x}_1 \dots \hat{x}_n)$, a similarity $\mathrm{sim}$ and an acceptance threshold $\theta$, the match set is the one-to-one partial assignment maximising total similarity, keeping only accepted pairs:
\begin{equation}
\begin{aligned}
  M_\theta(X,\hat{X}) = \big\{ (i,\pi(i)) :\ & i\in\mathrm{dom}(\pi),\\
  & \mathrm{sim}(x_i, \hat{x}_{\pi(i)}) \geq \theta \big\}
\end{aligned}
\label{eq:match}
\end{equation}
\begin{equation}
  \pi = \arg\max_{\pi \in \Pi} \sum_{i\in\mathrm{dom}(\pi)} \mathrm{sim}(x_i, \hat{x}_{\pi(i)}) ,
  \label{eq:hungarian}
\end{equation}
solved by Hungarian assignment after padding the rectangular similarity matrix with zero-weight dummy items; $\Pi$ is the resulting set of one-to-one partial maps. Precision, recall and $F_1$ of any matched family follow as
\begin{equation}
  P = \frac{|M_\theta|}{n}, \quad
  R = \frac{|M_\theta|}{m}, \quad
  F_1 = \frac{2PR}{P+R} ,
  \label{eq:prf}
\end{equation}
with a zero denominator defined to give score zero (and an empty gold--empty prediction pair defined as one).

\paragraph{Sections.} Section items are normalised titles, $\mathrm{sim}$ is normalised-title similarity, $\theta = 0.8$, with a positional tie-breaker for repeated titles; \textbf{Sec~P/R/F1} are Eq.~\ref{eq:prf}. 
Writing $M = M_{0.8}$ for the matched sections and $(i, \pi(i))$ for their gold and predicted positions, section order is Kendall's $\tau_b$ over matched pairs only,
\begin{equation}
  \mathrm{Sec}\,\tau = \frac{C - D}{\sqrt{(n_0 - n_1)(n_0 - n_2)}} ,
  \label{eq:tau}
\end{equation}
with $C$ and $D$ the concordant and discordant pairs of $M$, $n_0 = \binom{|M|}{2}$ and $n_1, n_2$ the tie corrections on each side. 
A missing section is therefore charged to detection, not to order; $\tau_b$ is undefined for $|M| < 2$ and those documents are excluded from the mean and counted.

\paragraph{Paragraphs.} Paragraph items are matched on the similarity of their first 200 normalised characters at $\theta = 0.7$, giving $M^p$; \textbf{Para~P/R} are Eq.~\ref{eq:prf} over $M^p$. 
A matched paragraph is \emph{correctly placed} iff the section containing its prediction is itself the match of the section containing its gold, so
\begin{equation}
  \mathrm{P}\!\to\!\mathrm{S} = \frac{1}{|M^p|} \sum_{(i,j) \in M^p} \mathbf{1}\big[\,(\sigma(i), \hat{\sigma}(j)) \in M \,\big] ,
  \label{eq:p2s}
\end{equation}
with $\sigma(i)$ and $\hat{\sigma}(j)$ the gold and predicted sections of the two paragraphs. P$\to$S conditions on matched paragraphs and must be read together with Para~P/R. Within-section order W$\tau$ is Eq.~\ref{eq:tau} applied to the correctly placed paragraphs of each matched section, averaged over sections weighted by their paragraph count.

\paragraph{Figures.} Gold \texttt{<fig>} captions are matched to predicted TEI \texttt{<figDesc>} strings at $\theta = 0.7$ (Eq.~\ref{eq:match}), excluding tables. Recall uses Eq.~\ref{eq:prf}, but the precision denominator is \emph{every} serialised non-table \texttt{<figure>}, captionless ones included:
\begin{equation}
  P_{\mathrm{fig}} = \frac{|M^f|}{|\hat{F}|}, \quad
  R_{\mathrm{fig}} = \frac{|M^f|}{|F|} ,
  \label{eq:fig}
\end{equation}
where $F$ are the gold figures and $\hat{F}$ all emitted non-table figures. Conditioning $\hat{F}$ on caption-bearing output instead would make a captionless region neither true nor false positive (\S\ref{sec:figures}).

\paragraph{Tables.} Table scores are those of the external benchmark \citep{soric2025benchmarking}: detection $F_1$ counts a predicted table region as matched at $\mathrm{IoU} > 0.5$ and applies Eq.~\ref{eq:prf}. 
For a gold grid $T$ and prediction $\hat{T}$ decomposed into cells, GriTS \citep{grits} is the two-dimensional generalisation of the longest-common-subsequence $F$-score over the maximal aligned substructures $\mathcal{A}$,
\begin{equation}
  \mathrm{GriTS}(T,\hat{T}) = \frac{2 \sum_{(c,\hat{c}) \in \mathcal{A}} f(c, \hat{c})}{|T| + |\hat{T}|} ,
  \label{eq:grits}
\end{equation}
with $f$ comparing row/column spans for GriTS\textsubscript{Top} and cell text for GriTS\textsubscript{Con}. TEDS \citep{teds} compares the two tables as HTML trees,
\begin{equation}
  \mathrm{TEDS}(T,\hat{T}) = 1 - \frac{\mathrm{EditDist}(T, \hat{T})}{\max(|T|, |\hat{T}|)} ,
  \label{eq:teds}
\end{equation}
with $\mathrm{EditDist}$ the tree edit distance and $|\cdot|$ the node count. 
GriTS and TEDS are computed over \emph{detected} tables only, so they must be read together with detection $F_1$.

\paragraph{Validation.} Table~\ref{tab:perturb} scores 200 perturbed copies of the gold reference against the original. Each perturbation moves exactly the score built to detect it (bold) and leaves the others at identity, including the injected caption, the leakage error most characteristic of layout failure.

\begin{table}[h]
\centering
\scriptsize
\setlength{\tabcolsep}{2.6pt}
\caption{Perturbation validation on 200 gold documents (mean score of perturbed vs.\ original; identity $=1.0$ everywhere). W$\tau$ = within-section order.}
\label{tab:perturb}
\begin{tabular}{lcccccc}
\toprule
Perturbation & Sec F1 & Sec $\tau$ & P$\to$S & W$\tau$ & Para P & Para R  \\
\midrule
swap adjacent paras & 1 & 1 & 1 & \textbf{.89} & 1 & 1 \\
move para across sec. & 1 & 1 & \textbf{.97} & 1 & 1 & 1 \\
merge two sections & \textbf{.95} & 1 & 1 & 1 & 1 & 1 \\
inject caption para & 1 & 1 & 1 & 1 & \textbf{.97} & 1 \\
drop a paragraph & 1 & 1 & 1 & 1 & 1 & \textbf{.97} \\
swap two sections & 1 & \textbf{.95} & 1 & 1 & 1 & 1 \\
\bottomrule
\end{tabular}
\end{table}

\section{Figure Region Granularity}
\label{app:figures}

Two merging strategies were compared: parent-anchored, where each box takes the nearest free caption, and caption-anchored (\S\ref{sec:pipeline}).

\begin{table}[h]
\centering
\small
\setlength{\tabcolsep}{4.5pt}
\begin{tabular}{llcccc}
\toprule
Corpus & System & P & R & F1 & Cap. \\
\midrule
\multirow{2}{*}{LS}
 & GROBID+PP & 0.709 & \textbf{0.855} & \textbf{0.775} & \textbf{0.912} \\
 & GROBID    & \textbf{0.710} & 0.726 & 0.718 & 0.903 \\
\midrule
\multirow{2}{*}{MS}
 & GROBID+PP & \textbf{0.813} & \textbf{0.971} & \textbf{0.885} & \textbf{0.956} \\
 & GROBID    & 0.799 & 0.927 & 0.858 & 0.934 \\
\bottomrule
\end{tabular}
\caption{Caption-linked figure recovery on Bioinformatics (LS) and Materials Science (MS). 
Gold \texttt{<fig>} captions matched one-to-one to TEI \texttt{<figDesc>} at $\theta = 0.70$. 
Cap.\ is caption fidelity on matched pairs. 
Best per corpus in \textbf{bold}.}
\label{tab:figures-body}
\end{table}

Under parent-anchored merging, the panels that do not own their figure's caption serialise as captionless \texttt{<figure>} elements: $57\%$ of GROBID+PP's Bioinformatics figures and $42\%$ on Materials Science. 
Anchoring the merge on the caption instead (\S\ref{sec:pipeline}) reduces the mean absolute error in region count per document from $6.13$ to $0.70$ on Materials Science, and makes the count exactly right on $65\%$ of its documents against $6\%$ under parent-anchored merging. 
One region per logical figure is the right target: DocLayNet's scientific-article pages carry $0.97$ \emph{Picture} annotations per \emph{Caption}, as does JATS.

\section{Table Extraction, Full Results}
\label{app:tables}

The full results from the table extraction are reported in Table~\ref{tab:table-full}.

\begin{table*}[h]
\centering
\footnotesize
\setlength{\tabcolsep}{3pt}
\caption{Table extraction under the protocol of \citet{soric2025benchmarking}, both cell-matching modes. \emph{bbox}: cells matched by bounding-box IoU; \emph{token}: matched by text tokens. F1 is table detection at IoU~$>0.5$; G$_{\mathrm{T}}$ and G$_{\mathrm{C}}$ are GriTS\textsubscript{Top} (grid topology) and GriTS\textsubscript{Con} (cell content). Plain GROBID and Docling rows are taken from the benchmark's released evaluation files, \texttt{GROBID+PP} is evaluated under the same protocol. Best per dataset and mode in \textbf{bold}.}
\label{tab:table-full}
\begin{tabular}{ll cccc cccc}
\toprule
& & \multicolumn{4}{c}{\textbf{bbox} mode} & \multicolumn{4}{c}{\textbf{token} mode} \\
\cmidrule(lr){3-6}\cmidrule(lr){7-10}
Dataset & System & F1 & G$_{\mathrm{T}}$ & G$_{\mathrm{C}}$ & TEDS & F1 & G$_{\mathrm{T}}$ & G$_{\mathrm{C}}$ & TEDS \\
\midrule
\multirow{3}{*}{PubTables}
 & GROBID+PP & 0.975 & 0.884 & 0.763 & 0.715 & 0.904 & 0.886 & 0.769 & 0.721 \\
 & GROBID    & 0.666 & 0.774 & 0.658 & 0.605 & 0.606 & 0.783 & 0.671 & 0.616 \\
 & Docling   & \textbf{0.988} & \textbf{0.952} & \textbf{0.866} & \textbf{0.858} & \textbf{0.962} & \textbf{0.958} & \textbf{0.878} & \textbf{0.870} \\
\midrule
\multirow{3}{*}{arXiv}
 & GROBID+PP & \textbf{0.926} & \textbf{0.862} & 0.662 & 0.583 & 0.572 & \textbf{0.865} & 0.669 & 0.589 \\
 & GROBID    & 0.430 & 0.781 & 0.617 & 0.535 & 0.342 & 0.802 & 0.646 & 0.562 \\
 & Docling   & 0.897 & 0.794 & \textbf{0.686} & \textbf{0.656} & \textbf{0.696} & 0.796 & \textbf{0.697} & \textbf{0.666} \\
\midrule
\multirow{3}{*}{BRGM}
 & GROBID+PP & \textbf{0.937} & 0.777 & 0.679 & 0.602 & 0.703 & 0.777 & 0.681 & 0.604 \\
 & GROBID    & 0.160 & 0.265 & 0.206 & 0.159 & 0.096 & 0.265 & 0.205 & 0.159 \\
 & Docling   & 0.906 & \textbf{0.823} & \textbf{0.738} & \textbf{0.739} & \textbf{0.824} & \textbf{0.827} & \textbf{0.744} & \textbf{0.745} \\
\midrule
\multirow{3}{*}{ICDAR}
 & GROBID+PP & 0.948 & 0.826 & 0.733 & 0.687 & 0.755 & 0.826 & 0.741 & 0.691 \\
 & GROBID    & 0.157 & 0.592 & 0.514 & 0.482 & 0.124 & 0.593 & 0.514 & 0.483 \\
 & Docling   & \textbf{0.990} & \textbf{0.969} & \textbf{0.896} & \textbf{0.895} & \textbf{0.990} & \textbf{0.983} & \textbf{0.940} & \textbf{0.938} \\
\bottomrule
\end{tabular}
\end{table*}

\section{Cost Model}
\label{app:cost}

The CPU rows of Table~\ref{tab:cost} are measured wall-time priced at Modal's published rates. 
With $t$ the pipeline time per document, $c{=}10$ reserved cores at $r_{\mathrm{cpu}}{=}\$1.31{\times}10^{-5}$/core\,s and $M{=}16$~GiB at $r_{\mathrm{mem}}{=}\$2.22{\times}10^{-6}$/GiB\,s, the cost per document is 
\begin{equation}
C_{\mathrm{doc}} = t\,(c\,r_{\mathrm{cpu}} + M\,r_{\mathrm{mem}}).
\label{eq:cost}
\end{equation}
Pricing the 2016-era Xeon E5-2650 v4 at Modal's current rate is conservative for the CPU rows.
The GPU rows instead divide the total billed \texttt{modal.com} spend for the campaign by the documents produced, so they include container start-up, model loading and idle time until scale-down, which a per-request timing would miss.
That share differs by deployment, and the \texttt{Docling} endpoint also served several requests per container on whichever GPU type was free, so the per-second rate implied by the s/doc and \$/doc columns is not the same across systems: \texttt{Docling}'s billed seconds per document fall below its request latency, the generative endpoints' exceed it.
The rows are what each system cost to run, not a normalised hourly rate.
The \texttt{MinerU} row is a 100-document sample rather than a campaign, so container start-up and warm idle time are a larger share of its bill than of the others'.

\section{Docling on CPU}
\label{app:docling-cpu}
\texttt{Docling}'s standard pipeline also runs without a GPU, so its CPU cost is the natural check on the margin of \S\ref{sec:cost}.
We timed the pipeline under the conditions of the Modal endpoint: the same standard pipeline and threaded docling-parse backend that \texttt{docling-serve} 1.7.0 selects by default, with the request options of the evaluation (OCR on, TableFormer in accurate mode, images as placeholders), the only difference being the library version, 2.126 here against the 2.38-or-later series the endpoint image pins.
The machine is the 12-core Xeon E5-2650 v4 of Table~\ref{tab:cost}; the samples are 20 random documents per corpus, two worker processes sharing the twelve cores ($12/w$ threads each).
Cost is Eq.~\ref{eq:cost} with $c{=}12$.

At that best setting (two workers) \texttt{Docling} on CPU takes $49.3$~s per Materials Science and $56.1$~s per Bioinformatics document, six and ten times \texttt{GROBID+PP} on the same cores, and at the tariff of Table~\ref{tab:cost} costs \$9,500 and \$10,800 per million documents against \$1,345 and \$968, seven to eleven times more, and two and a half to three and a half times more than its own GPU endpoint (\$3,645 and \$3,090).
The GPU figure in Table~\ref{tab:cost} is therefore \texttt{Docling}'s cheaper configuration, and the margin of \S\ref{sec:cost} is against it.

\paragraph{Caveats.} These are estimates, not a campaign: 10 to 20 documents per configuration against the full corpora behind Table~\ref{tab:cost}, and one machine.
The runs are pinned to twelve cores and priced at twelve reserved cores rather than the ten of the \texttt{GROBID+PP} rows, which favours \texttt{GROBID+PP} by 17\% in the cost ratio and not at all in the time ratio.
Model loading (40--71~s) is excluded, as it is for every CPU row.
No attempt was made to tune \texttt{Docling} for CPU beyond the worker sweep; a build with OCR disabled would be faster on born-digital PDFs but would not be the configuration that was evaluated.

\section{OmniDocBench, Scientific-Article Subset}
\label{app:omnidocbench}

\texttt{OmniDocBench} scores each page in isolation, whereas \texttt{GROBID}'s models label one token stream carrying whole-article structure (title block, section sequence, reference list), so a page taken from mid-article gives them nothing to condition on; the extension inherits the constraint, since masking changes only which tokens enter that stream.
The distribution compounds it: v1.5 and v1.6 ship page images and \texttt{GROBID} performs no OCR, so the released benchmark carries no font stream at all, and over the 981 single-page PDFs of the archived \texttt{v1\_0} branch \texttt{GROBID} returns no body text for $\approx\!44\%$ of pages (handwritten notes $100\%$, slides $80\%$, textbooks $70\%$).
The benchmark therefore cannot be run whole against a document-level parser, hence the \emph{n/a} in Table~\ref{tab:taxonomy}, but the part of it that is well posed for one can. We therefore evaluate on the intersection of two conditions: v1.5 annotations for pages whose \texttt{data\_source} is \texttt{academic\_literature}, and for which a source PDF is distributed. Only the archived \texttt{v1\_0} branch ships PDFs, so this yields \textbf{129 of the 215} v1.5 academic pages; the remaining 86 were added in v1.5 and exist as images only. All 129 are English and all carry a text layer (median 3{,}829 characters). The ground truth over them holds 695 text blocks, 161 titles, 260 references, 94 figures, 79 tables and 48 display formulas.

Both arms use the same \texttt{GROBID} build, the same request payload and the same TEI-to-Markdown conversion, so the only difference between them is the mask; the detector, merging strategy and parameters are those of \S\ref{sec:pipeline}. 

\begin{table}[H]
\centering
\small
\setlength{\tabcolsep}{4pt}
\begin{tabular}{@{}lccc@{}}
\toprule
Metric & & GROBID & GROBID+PP \\
\midrule
TextEdit        & $\downarrow$ & \textbf{0.116} & 0.128 \\
TableEdit       & $\downarrow$ & 0.581 & \textbf{0.319} \\
TableTEDS       & $\uparrow$   & 31.47 & \textbf{54.75} \\
TableTEDS-S     & $\uparrow$   & 36.45 & \textbf{63.49} \\
ReadOrderEdit   & $\downarrow$ & 0.267 & \textbf{0.261} \\
\midrule
Tables emitted  &              & 46 & \textbf{75} \\
\bottomrule
\end{tabular}
\caption{\texttt{OmniDocBench} v1.5 end-to-end metrics on the 129-page scientific-article subset, \texttt{quick\_match}. \emph{Tables emitted} is what each arm produced, against 79 tables in the ground truth.}
\label{tab:omnidocbench-subset}
\end{table}

Table~\ref{tab:omnidocbench-subset} reproduces the table-structure result of \S\ref{sec:tables} on a benchmark this work had not otherwise used: TEDS rises from $31.5$ to $54.7$ and structure-only TEDS from $36.5$ to $63.5$, with tables emitted going from 46 to 75 against 79 in the ground truth. As in \S\ref{sec:tables}, the table model is unchanged and is simply handed a bounded region.

Text edit distance moves the other way by $0.012$, and the metric's construction accounts for it: \texttt{OmniDocBench} filters \texttt{figure\_caption}, \texttt{table\_caption}, the footnote categories, \texttt{header}, \texttt{footer}, \texttt{page\_footnote}, \texttt{page\_number} and \texttt{equation\_caption} out of the ground truth before scoring text blocks. Those are exactly the categories the masking suppresses, so the benefit of paratext masking is invisible to this metric by construction while its marginal cost still registers. This is a difference in what the two protocols measure, not a reversal of the normalised-similarity gains of \S\ref{sec:body}. Display-formula metrics are not reported: \texttt{GROBID} emits the raw text of an equation rather than \LaTeX{}, so an edit distance against \LaTeX{} ground truth measures that representation mismatch rather than anything the mask changes.

Two limits apply to these numbers. Each page is still scored in isolation, so both arms are handicapped equally by the missing document context, and neither figure is \texttt{GROBID}'s behaviour on whole articles. And no vision-based system has been run on this subset, so Table~\ref{tab:omnidocbench-subset} supports the paired within-parser comparison only, and none of its values may be read against the published leaderboard, which scores the full 1{,}651-page set.

\end{document}